\pdfoutput=1

\documentclass[11pt]{article}

\usepackage[final]{acl}

\usepackage{times}
\usepackage{latexsym}

\usepackage[T1]{fontenc}

\usepackage[utf8]{inputenc}

\usepackage{microtype}

\usepackage{inconsolata}

\usepackage{maltese}

\usepackage{graphicx}

\usepackage{enumitem}

\usepackage{multirow}
\usepackage{color, colortbl}

\usepackage{subcaption}

\title{MELABenchv1: Benchmarking Large Language Models against \\ Smaller Fine-Tuned Models for Low-Resource Maltese NLP}

\author{Kurt Micallef \\
  \texttt{kurt.micallef@um.edu.mt} \\\And
  Claudia Borg \\
  \texttt{claudia.borg@um.edu.mt} \\\AND
 {\normalfont Department of Artificial Intelligence, University of Malta}
}

\newcommand{\PT}{\texttt{PT}}
\newcommand{\IT}{\texttt{IT}}
\newcommand{\NO}{\texttt{NO}}
\newcommand{\NK}{\texttt{NK}}

\newcommand{\NLU}{discriminative}
\newcommand{\NLG}{generative}

\begin{document}
\maketitle

\begin{abstract}
Large Language Models (LLMs) have demonstrated remarkable performance across various Natural Language Processing (NLP) tasks, largely due to their generalisability and ability to perform tasks without additional training.
However, their effectiveness for low-resource languages remains limited. In this study, we evaluate the performance of 55 publicly available LLMs on Maltese, a low-resource language, using a newly introduced benchmark covering 11 \NLU{} and \NLG{} tasks.
Our experiments highlight that many models perform poorly, particularly on \NLG{} tasks, and that smaller fine-tuned models often perform better across all tasks.
From our multidimensional analysis, we investigate various factors impacting performance. We conclude that prior exposure to Maltese during pre-training and instruction-tuning emerges as the most important factor.
We also examine the trade-offs between fine-tuning and prompting, highlighting that while fine-tuning requires a higher initial cost, it yields better performance and lower inference costs.
Through this work, we aim to highlight the need for more inclusive language technologies and recommend that researchers working with low-resource languages consider more ``traditional'' language modelling approaches.
\end{abstract}

\section{Introduction}

Large Language Models (LLMs) have seen a huge rise in use due to their strong performance and remarkable generalisability across a wide range of diverse tasks \cite{openai-2024-gpt4, grattafiori-etal-2024-llama3, gemma-2024-gemma2}.
Their appeal is evident from their ability to perform tasks without additional training, with models able to infer from a few examples (few-shot or in-context learning) or an instruction (zero-shot) \cite{raffel-etal-2020-t5, brown-etal-2020-gpt3}.
Furthermore, the availability of a plethora of multilingual models makes this technology more accessible for many languages, due to the inherent cross-lingual transfer capabilities.
Despite this, many low-resource languages still face significant challenges in achieving strong performance with these models \cite{ahuja-etal-2023-mega, asai-etal-2024-buffet}.

Prior research on older multilingual models, such as mBERT, highlighted the challenges faced by low-resource languages, particularly when a language is absent from pre-training \cite{chau-etal-2020-parsing, muller-etal-2021-unseen}.
However, modern LLMs often have an additional training phase, designed to improve their generalisability: instruction-tuning.
This raises new questions about the extent to which instruction-tuning mitigates or exacerbates performance gaps for low-resource languages.

In this work, we aim to address this for Maltese, an official EU language that ranks the lowest in the Digital Language Equality score \cite{rosner-borg-2022-ele_maltese}.
While our primary objective is to understand which LLM properties influence downstream task performance, we compare this to more traditional fine-tuning approach with relatively smaller models.
The main contributions of this work are:

\begin{enumerate}\itemsep0em
    \item
    A new evaluation benchmark composed of a variety of 11 \NLU{} and \NLG{} Maltese NLP tasks, to facilitate the evaluation and development of language technology.
    \item
    A comprehensive experimental setup on 55 LLMs, for which we analyse which LLM properties are most important for better downstream task performance.
    \item
    Several fine-tuned models of a relatively smaller size for each of these tasks, often surpassing all LLMs included in this study.
\end{enumerate}

Our evaluation code and results are made publicly available.\footnote{
\url{https://huggingface.co/spaces/MLRS/MELABench}
}
We also make the best performing fine-tuned models publicly available.\footnote{
BERTu: \url{https://huggingface.co/collections/MLRS/bertu-683ac54c1b6ab3ae715cb43d};\\
mT5-Small: \url{https://huggingface.co/collections/MLRS/mt5-small-683eecd001179a722c98298b}.
}
Through our evaluation, we explore the following research questions:
\begin{enumerate}\itemsep0em
    \item
    How well do LLMs perform compared to smaller fine-tuned models?
    \item
    What factors contribute to a model's performance on downstream tasks?
    \item
    How viable is it to train smaller but task-specific models as opposed to prompting larger but generic models?
\end{enumerate}

\section{Experimental Setup}
\label{section:experiments}

\subsection{Evaluation Benchmark}
\label{section:tasks}

We conducted a survey of publicly available Maltese datasets, which allows us to benchmark the performance of various models on Maltese.
We make a distinction with the type of task depending on whether the output is discrete (\NLU{}) or a text in natural language (\NLG{}).
In total, we collected 11 datasets, shown in Table~\ref{table:datasets}, with an even mixture of \NLU{} and \NLG{} tasks.

\begin{table*}[t!]
    \centering
    \setlength{\tabcolsep}{5pt}
    \footnotesize
    \begin{tabular}{|c|l|l|r|r|r|}
        \hline
        \textbf{Type} & \textbf{Name} & \textbf{Task} & \multicolumn{1}{c|}{\textbf{$|$train$|$}} & \multicolumn{1}{c|}{\textbf{$|$validation$|$}} & \multicolumn{1}{c|}{\textbf{$|$test$|$}} \\
        \hline\hline
        \multirow{6}{*}{\rotatebox[origin=c]{90}{\NLU}} & Sentiment \cite{martinez-garcia-etal-2021-evaluating} & Sentiment Analysis & 595 & 85 & 433 \\
        & SIB-200 \cite{adelani-etal-2024-sib} & Topic Classification & 701 & 99 & 204 \\
        & Taxi1500 \cite{ma-etal-2024-taxi1500} & Topic Classification & 860 & 106 & 111 \\
        & News Categories \cite{chaudhary-etal-2024-topic} & Topic Classification (Multi-Label) & 10,784 & 2,293 & 2,297 \\
        & MultiEURLEX \cite{chalkidis-etal-2021-multieurlex} & Topic Classification (Multi-Label) & 17,521 & 5,000 & 5,000 \\
        & Belebele \cite{bandarkar-etal-2024-belebele} & Machine Reading Comprehension & 0 & 0 & 900 \\
        \hline
        \multirow{5}{*}{\rotatebox[origin=c]{90}{\NLG}} & OPUS-100 Fixed \cite{abela-etal-2024-tokenisation} & Machine Translation (EN$\rightarrow$MT) & 1,000,000 & 2,000 & 2,000 \\
        & Flores-200 \cite{nllb-2022-nllb} & Machine Translation (EN$\rightarrow$MT) & 0 & 997 & 1,012 \\
        & WebNLG \cite{cripwell-etal-2023-2023} & Data-to-Text & *13,211 & 1,665 & 1,778 \\
        & EUR-Lex-Sum \cite{aumiller-etal-2022-eur} & Abstractive Summarisation & 940 & 187 & 188 \\
        & News Headlines \cite{chaudhary-etal-2024-topic} & Abstractive Summarisation & 17,782 & 3,810 & 3,811 \\
        \hline
    \end{tabular}
    \caption{Dataset Summary}
    \label{table:datasets}
    {
        \footnotesize*Indicates noisy data obtained through machine translation.\\
    }
\end{table*}

\subsection{Models}
\label{section:models}

To answer our primary research question, we use a variety of generative models.
We consider 55 different language models whose weights are publicly available, covering various properties which we consider important for our analysis.
These are model size (300M -- 15B), language coverage (18 -- 511, where known), and whether the model is pre-trained (\PT{}) or instruction-tuned (\IT{}).
Moreover, we identify whether the model has seen Maltese during pre-training (\PT{}), during instruction-tuning (\IT{}), or never (\NO{}). In the case of commercially released models, this information is not available and is categorised as unknown (\NK{}).
These details are summarised in Table~\ref{table:models}.
Additionally, in Appendix~\ref{appendix:chatgpt_result} we present an evaluation for ChatGPT 4o in a limited experimental setup.

\definecolor{PT/NO}{HTML}{1F77B4}
\definecolor{IT/NO}{HTML}{FF7F0E}
\definecolor{PT/PT}{HTML}{2CA02C}
\definecolor{IT/PT}{HTML}{D62728}
\definecolor{IT/IT}{HTML}{9467BD}
\definecolor{PT/NK}{HTML}{8C564B}
\definecolor{IT/NK}{HTML}{E377C2}

\begin{table*}[t!]
    \centering
    \setlength{\tabcolsep}{5pt}
    \footnotesize
    \begin{tabular}{|l||l|r|c|}%
        \hline
        \textbf{Name} & \textbf{Parameter Count} & \textbf{Languages} & \textbf{Training} \\
        & & & \scriptsize{overall/Maltese} \\
        \hline\hline
        PolyLM \cite{wei-etal-2023-polylm} & 1.7B, 13B & 18 & \cellcolor{PT/NO!50} \PT/\NO \\
        XGLM \cite{lin-etal-2022-shot} & 564M, 1.7B, 2.9B, 4.5B, 7.5B & 30 & \cellcolor{PT/NO!50} \PT/\NO \\
        mGPT \cite{shliazhko-etal-2024-mgpt} & 1.3B, 13B & 61 & \cellcolor{PT/NO!50} \PT/\NO \\
        BLOOM \cite{bigscience-2023-bloom} & 560M, 2B, 3B, 8B & 46 & \cellcolor{PT/NO!50} \PT/\NO \\
        Aya-23 \cite{aryabumi-etal-2024-aya23} & 8B & 23 & \cellcolor{IT/NO!50} \IT/\NO \\
        BLOOMZ \cite{muennighoff-etal-2023-crosslingual} & 560M, 2B, 3B, 8B & 46 & \cellcolor{IT/NO!50} \IT/\NO \\
        BX-LLaMA \cite{li-etal-2023-bactrianx} & 7B, 13B & *52 & \cellcolor{IT/NO!50} \IT/\NO \\
        BX-BLOOM \cite{li-etal-2023-bactrianx} & 7B & *77 & \cellcolor{IT/NO!50} \IT/\NO \\
        \hline
        Salamandra \cite{gonzalez-agirre-etal-2025-salamandra} & 2B, 7B & 35 & \cellcolor{PT/PT!50} \PT/\PT \\
        EuroLLM \cite{matrins-etal-2025-eurollm} & 1.7B, 9B & 35 & \cellcolor{PT/PT!50} \PT/\PT \\
        mT5 \cite{xue-etal-2021-mt5} & 300M, 582M, 1.23B, 3.74B, 13B & 101 & \cellcolor{PT/PT!50} \PT/\PT \\
        MaLA-500 \cite{lin-etal-2024-mala500} & 8.6B & 511 & \cellcolor{PT/PT!50} \PT/\PT \\
        Teuken Instruct Research v0.4 \cite{ali-etal-2024-teuken} & 7B & *24 & \cellcolor{IT/PT!50} \IT/\PT \\
        Salamandra Instruct \cite{gonzalez-agirre-etal-2025-salamandra} & 2B, 7B & *35 & \cellcolor{IT/PT!50} \IT/\PT \\
        mT0 \cite{muennighoff-etal-2023-crosslingual} & 300M, 582M, 1.23B, 3.74B, 13B & *120 & \cellcolor{IT/PT!50} \IT/\PT \\
        \hline
        EuroLLM Instruct \cite{matrins-etal-2025-eurollm} & 1.7B, 9B & 35 & \cellcolor{IT/IT!50} \IT/\IT \\
        Aya-101 \cite{ustun-etal-2024-aya} & 13B & 101 & \cellcolor{IT/IT!50} \IT/\IT \\
        \hline
        Gemma 2 \cite{gemma-2024-gemma2} & 2B, 9B & ? & \cellcolor{PT/NK!50} \PT/\NK \\
        Llama 2 \cite{touvron-etal-2023-llama2} & 7B, 13B & ? & \cellcolor{PT/NK!50} \PT/\NK \\
        Llama 3 \cite{grattafiori-etal-2024-llama3} & 8B & ? & \cellcolor{PT/NK!50} \PT/\NK \\
        Ministral Instruct 2410 \cite{mistral-2024-ministral} & 8B & ? & \cellcolor{IT/NK!50} \IT/\NK \\
        Gemma 2 Instruct \cite{gemma-2024-gemma2} & 2B, 9B & ? & \cellcolor{IT/NK!50} \IT/\NK \\
        Llama 2 Chat \cite{touvron-etal-2023-llama2} & 7B, 13B & ? & \cellcolor{IT/NK!50} \IT/\NK \\
        Llama 3 Instruct \cite{grattafiori-etal-2024-llama3} & 8B & ? & \cellcolor{IT/NK!50} \IT/\NK \\
        \hline
    \end{tabular}
    \caption{Language Model Summary}
    \label{table:models}
    {
        *Since the set of languages used during \PT{} and \IT{} is not the same, the union of both sets is represented.
        \\
        ? = For models with closed-source data, the set of languages used during training is unknown.
    }
\end{table*}

\subsection{Evaluation}
\label{section:evaluation}

We use the Language Model Evaluation Harness \cite{lm-evaluation-harness} to conduct the prompting experiments.
For each task, we define a template in which we structure the input in textual format together with an instruction, as well as formatting the target in textual format.
For \NLG{} tasks, the output is simply given as is, but for \NLU{} tasks, discrete label(s) are mapped into textual format as necessary.
Our main experiments are conducted with English instructions, but we also include a set of experiments with Maltese instructions which we manually translate.
See Appendix~\ref{appendix:prompts} for further details regarding the prompt templates used.

In our setup, we conduct two main experiments: zero-shot and one-shot.
In the zero-shot case, the model is given only the input and the instruction, and it is expected to produce the corresponding output.
In the one-shot case, we additionally prepend this with the input and output of a sample from the given task, formatted with the same template.
In both cases, inference is carried out on the final instance, where the output is not provided to the model.
Any examples used for in-context learning (one-shot) are taken from the training set, when this is available, or the validation set otherwise.
Since no training or validation set is available for Belebele, this task is omitted from one-shot experiments.

In terms of automated evaluation metrics, we report the following.
We use macro-averaged F1 for Sentiment Analysis, SIB-200, Taxi1500, Maltese News Categories, and MultiEURLEX.
For Belebele, we report the accuracy.
We report ChrF scores for OPUS-100, Flores-200, and WebNLG, and Rouge-L for EUR-Lex-Sum and Maltese News Headlines.
Additionally, we also provide BLEU scores for OPUS-100 and Flores-200, and Rouge-L scores for WebNLG, and ChrF scores for EUR-Lex-Sum and Maltese News Headlines in Appendix~\ref{appendix:all_results}.

When evaluating the output, the appropriate metrics are calculated on the generated output for \NLG{} tasks.
For \NLU{} tasks, this is not as straightforward since the expected output is discrete.
Hence, the output is extracted by comparing the log-likelihood of generating each label.
The label with the highest log-likelihood is chosen for single-label classification tasks (Sentiment, SIB-200, Taxi1500, and Belebele).
For multi-label classification tasks (News Categories and MultiEURLEX), we extract the predicted labels based on the number of gold labels.

\subsection{Fine-Tuned Models}
\label{section:baselines}

We want to compare LLMs to the performance of smaller fine-tuned models, which also serve as baselines.
The models are trained on the training set by performing parameter updates on the model.
For each dataset, we train a separate model. 
Since no training set is available for Belebele and Flores-200, no baselines are fine-tuned for these tasks.

We consider BERT-based \PT{} models, for which we add a linear classification head on top of the \PT{} model.
These are BERTu \cite{micallef-etal-2022-pre} -- a monolingual Maltese 126M parameter model -- and mBERT \cite{devlin-etal-2019-bert} -- a multilingual model with 179M parameters.
However, these are only applied to \NLU{} tasks since they are encoder-only models.

Therefore, we also fine-tune mT5-Small \cite{xue-etal-2021-mt5} -- a multilingual encoder-decoder \PT{} model with 300M parameters.
Similar to the prompted models, we convert every input and output into textual format.
However, we simply train on the textual input-output pairs and do not apply any prompt templates.
Moreover, we do not include any task prefix which were used to fine-tune the original models \cite{raffel-etal-2020-t5, xue-etal-2021-mt5}.\footnote{
This decision was made because the model is fine-tuned separately on each task, and the prefix did not have much influence on performance during our initial tests.
}
Evaluation metrics for fine-tuning mT5 are otherwise computed similarly to prompted models.
More details on our fine-tuning setup are included in Appendix~\ref{appendix:experiments_finetuning}.

\section{Overall Trends}
\label{section:overall_analysis}

\begin{figure}[t!]
    \centering
    \includegraphics[width=\linewidth]{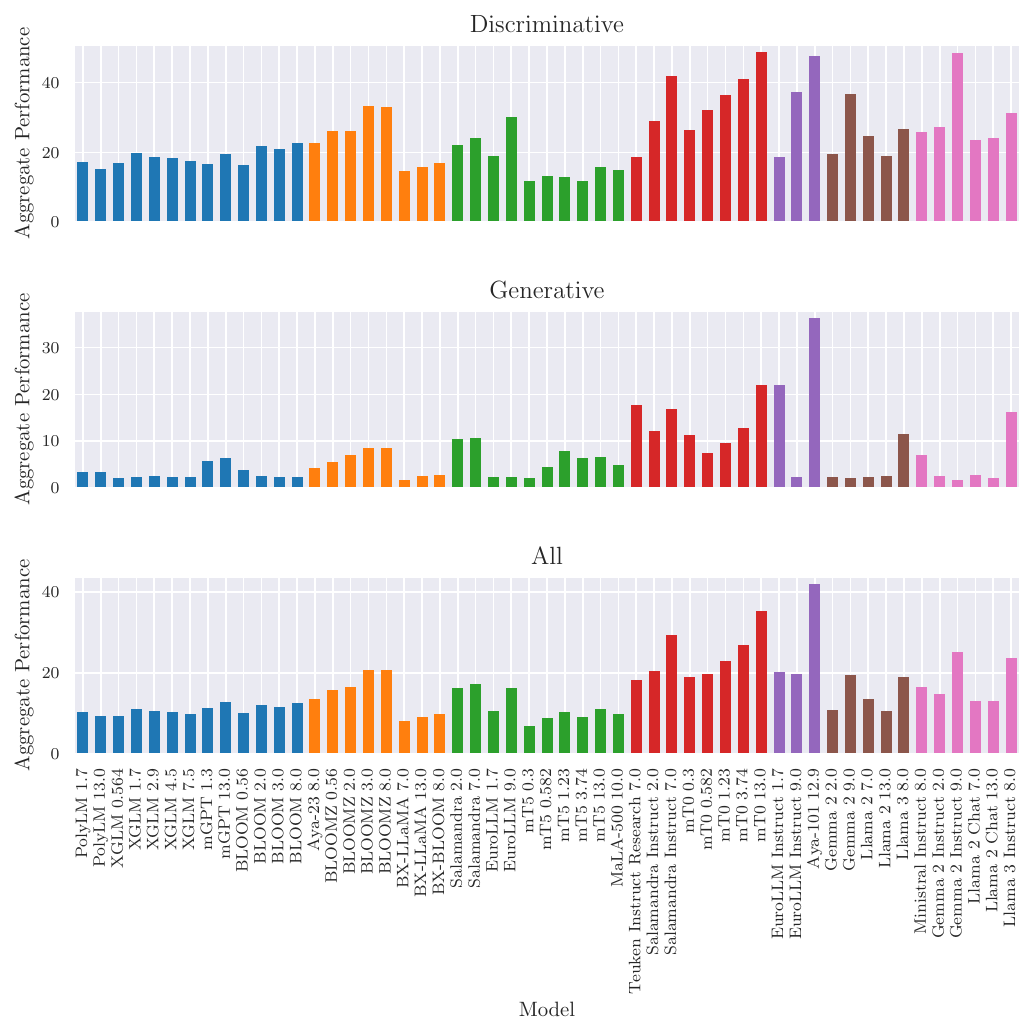}
    \caption{Zero-shot prompting performance of individual models aggregated across tasks.}
    \label{figure:model_0shot_performance}
\end{figure}

The starting point of our analysis is to understand the performance of all individual models across tasks by performing zero-shot prompting.
Thus, we aggregate the scores of each model averaged across \NLU{} and \NLG{} tasks.\footnote{
Although metrics in different tasks are not the same, we note that they are already normalised within the same range.
}
For each model, we then calculate an overall score across all tasks by averaging these two scores.
The results are shown in Figure~\ref{figure:model_0shot_performance}.

Overall, Aya-101 is the best-performing model, followed closely by mT0-XXL.
We attribute this primarily to the models' exposure to Maltese data, which we further investigate in Section~\ref{section:maltese_training}.
The smaller mT0 models are the next best-performing models overall, along with Salamandra Instruct 7B, Gemma 2 Instruct 9B, and Llama 3 Instruct 8B.

For \NLG{} tasks, Aya-101 performs better than any other model, often by a significant margin.
This is followed by mT0-XXL, EuroLLM Instruct 1.7B, Teuken Research Instruct 7B, Salamandra Instruct 7B, and Llama 3 Instruct 8B.
In the case of \NLU{} tasks, on average, mT0-XXL and Gemma 2 Instruct 9B perform better than Aya-101.

\begin{figure*}[t!]
    \centering
    \includegraphics[scale=0.5]{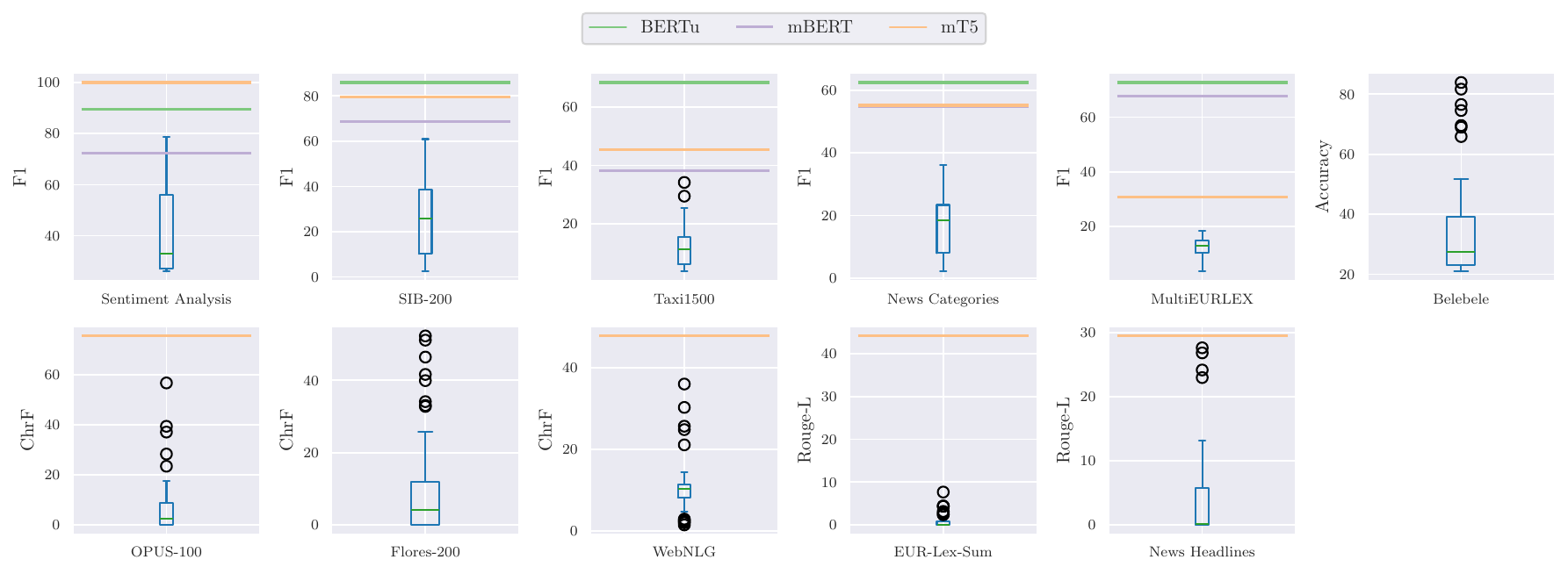}
    \caption{Zero-shot prompting performance distribution of models per task.}
    \label{figure:task_0shot_performance}
\end{figure*}

\begin{figure*}[t!]
    \centering
    \includegraphics[scale=0.5]{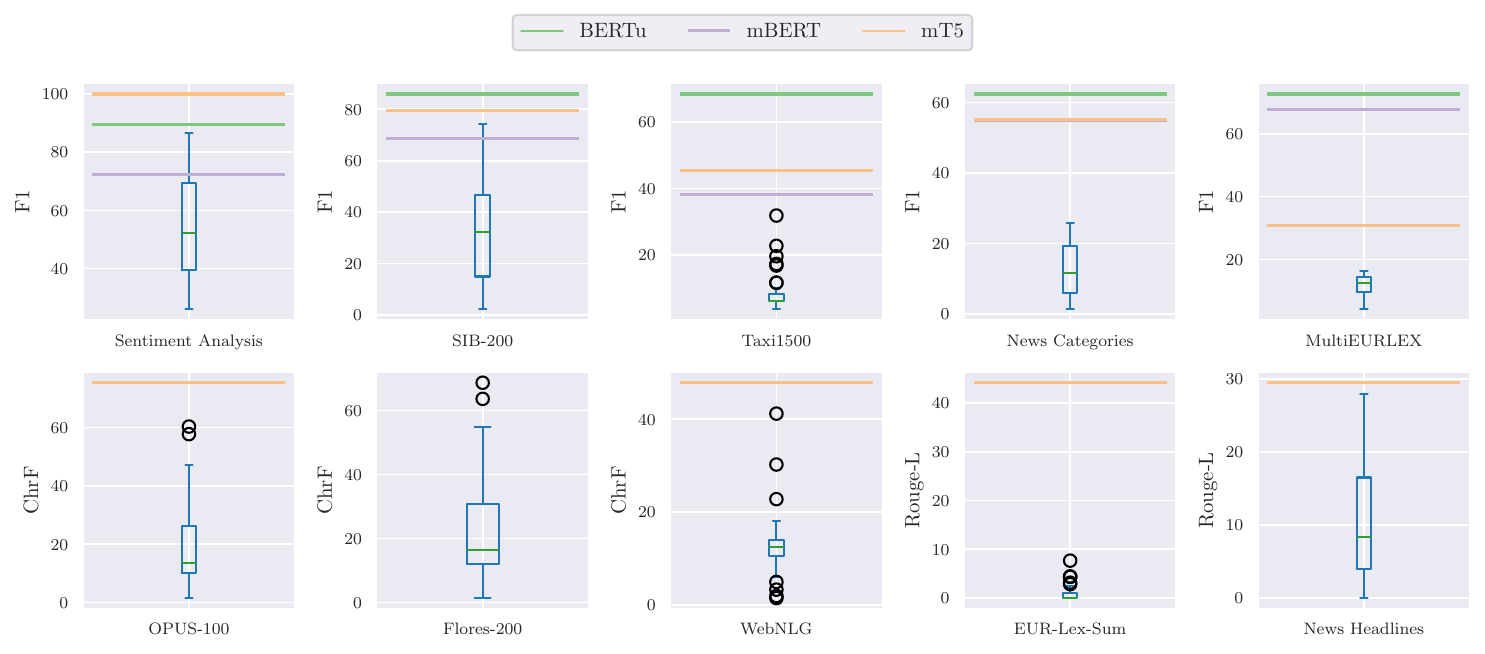}
    \caption{One-shot prompting performance distribution of models per task.}
    \label{figure:task_1shot_performance}
\end{figure*}

We take a closer look at the score distribution for each task in Figure~\ref{figure:task_0shot_performance}.
More models are competitive with the fine-tuned baselines on \NLU{} tasks than \NLG{} tasks.
Performance on \NLG{} tasks hovers near 0 for many models.
A more in-depth analysis of the individual models on each task (see Figure~\ref{figure:model_task_0shot_performance}) reveals that for \NLG{} tasks, the top performers are Aya-101, mT0-XXL, and Teuken Instruct Research, and to a lesser degree, EuroLLM 1.7B and Llama 3 Instruct 8B.
These models often act as outliers from the rest of the models.
The insights highlight that generating text data is much more challenging than understanding it and that many models fail to capture the linguistic nuances of a low-resource language like Maltese.

Looking at each task in Figure~\ref{figure:task_0shot_performance}, we observe that models generally struggle with Taxi1500 and MultiEURLEX among \NLU{} tasks and EUR-Lex-Sum among \NLG{} tasks.
This could be attributed to the specific domain of these tasks: the Bible for Taxi1500 and European Union documents for MultiEURLEX and EUR-Lex-Sum.
The latter two tasks have input sequences that are also generally longer, which hampers performance due to the model's limited context length.
Moreover, we note that MultiEURLEX is the only \NLU{} task where models perform quite on par with one another, particularly since it is a multi-label classification task on a massive scale.

When compared to the baselines, all prompted models perform worse, with the exception of the Sentiment Analysis task, for which some models outperform mBERT.
Among the baselines, mT5 generally performs better than mBERT, except for MultiEURLEX, potentially due to the task being ill-suited for generative models, as discussed earlier.
Overall, BERTu performs the best in \NLU{} tasks due to its Maltese pre-training, except for Sentiment Analysis for which we observe a perfect score for mT5.
While we are uncertain why the model does so well, we posit that this is due to the task being quite simplistic, since it is the only task where prompted models outperform some of the baselines.
Moreover, out of all the tasks, this dataset is the most likely to contain code-switching, which might make it easier for multilingual models to pick up on certain linguistic signals.

\subsection{Model Training}
\label{section:model_training}

\begin{figure*}[h!]
    \centering
    \includegraphics[scale=0.5]{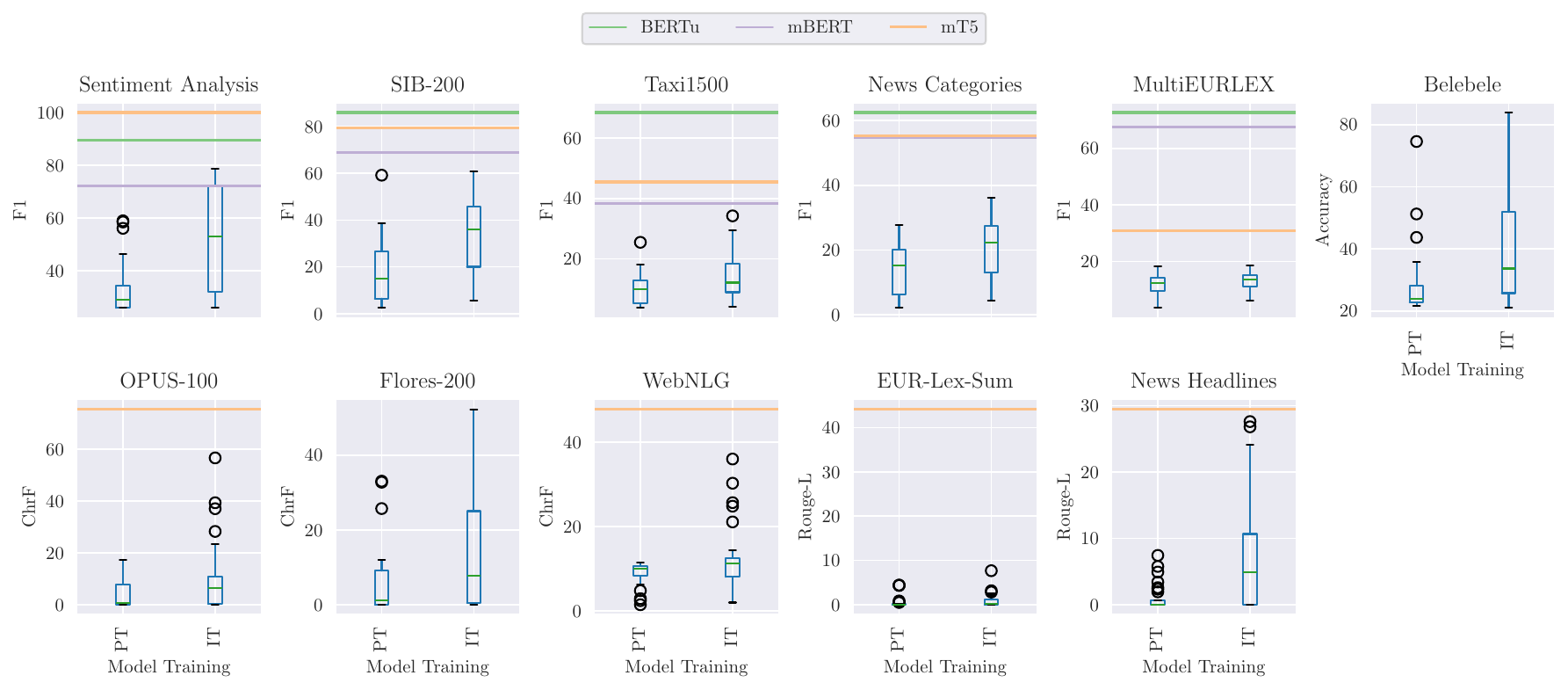}
    \caption{Zero-shot prompting performance distribution per task, with models grouped by different training types.}
    \label{figure:model_training_performance}
\end{figure*}

We now examine the relationship between zero-shot prompting performance and overall model training (\PT{} vs \IT{}).
We group models by their training and visualise the average performance for each task in Figure~\ref{figure:model_training_performance}.

It is very evident that \IT{} models perform better than \PT{} models on all tasks.
The performance difference between model types is quite small for some tasks, such as MultiEURLEX and EUR-Lex-Sum, which further reinforces our argument that these datasets are challenging for the prompted models.
The performance disparity on some tasks is proportionately larger, which could be due to the difficulty of the task and/or previous instruction-tuning on the same task.

\subsection{Number of Shots}
\label{section:shots}

We want to understand the potential impact of in-context learning.
We perform one-shot prompting and compare it to the previous zero-shot results.

Figure~\ref{figure:task_1shot_performance} shows the distribution of models on each of the tasks in relation to the baseline models with one-shot prompting.
Overall, the performance gap between fine-tuned models and prompted models is reduced.
In Sentiment Analysis, the best-performing prompted models perform almost as well as a fine-tuned BERTu and almost as good as a fine-tuned mT5-Small on Maltese News Headlines.
We also observe that some models perform better than fine-tuned mBERT on SIB-200 with one-shot compared to zero-shot.
However, for most of the other tasks, the gap between fine-tuned and prompted models remains significant.

\begin{figure}[t!]
    \centering
    \includegraphics[width=\linewidth]{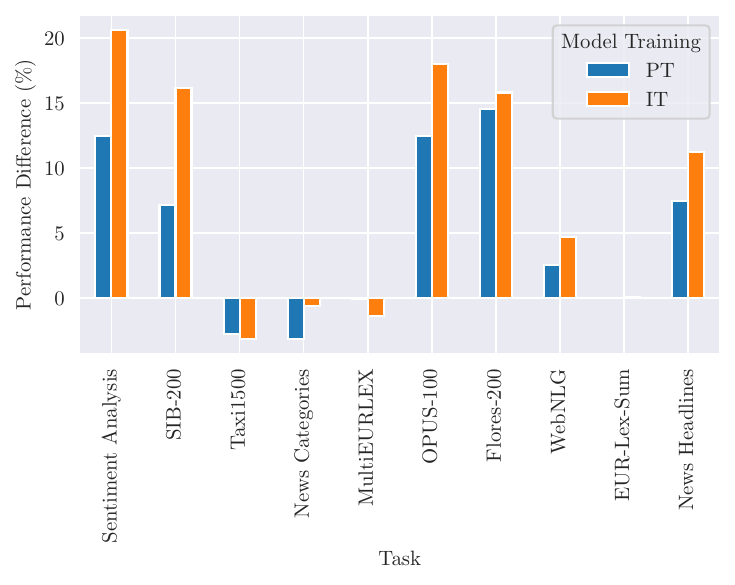}
    \caption{Performance difference of zero-shot prompting and one-shot prompting averaged across models with different training types.}
    \label{figure:shot_performance_difference}
\end{figure}

To better interpret the changes between zero-shot and one-shot, we calculate the performance difference aggregated across \PT{} and \IT{} models for each task.
This is calculated by subtracting the zero-shot performance from the corresponding one-shot performance of every prompted model.\footnote{
Similar to \citet{zhang-etal-2024-impact}, we observe negative effects for mT0 and BLOOMZ due to their zero-shot instruction-tuning, and hence we omit them for this analysis.
}
Figure~\ref{figure:shot_performance_difference} shows the performance difference, with a positive score indicating better one-shot results on average.

Similar to \citet{zhang-etal-2024-impact}, we observe consistent performance improvements with one-shot across all \NLG{} tasks, but mixed results in \NLU{} tasks, with slight degradations for Taxi1500, News Categories, and MultiEURLEX.
\IT{} models also give significant improvements in Sentiment Analysis, SIB-200, OPUS-100, and Maltese News Headlines.
The performance on MultiEURLEX and EUR-Lex-Sum is largely the same regardless of the number of shots.
We attribute this to the longer inputs which are being truncated to a limited sequence length.

\section{Effect of Maltese Exposure}
\label{section:maltese_analysis}

We now examine the effect that exposing the model to Maltese has on its performance.
Apart from a model's overall training (\PT{} or \IT{}), models are further grouped into different categories based on their explicit training on Maltese data, indicated by: \NO{}, \PT{}, and \IT{}, referring to no exposure to Maltese data, exposure during pre-training, and exposure during instruction-tuning, respectively.
For these analyses, we exclude models for which this information cannot be inferred from publicly available metadata (\NK{}).

\subsection{Maltese Training}
\label{section:maltese_training}

\begin{figure*}[t!]
\begin{subfigure}{\linewidth}
    \centering
    \includegraphics[scale=0.5]{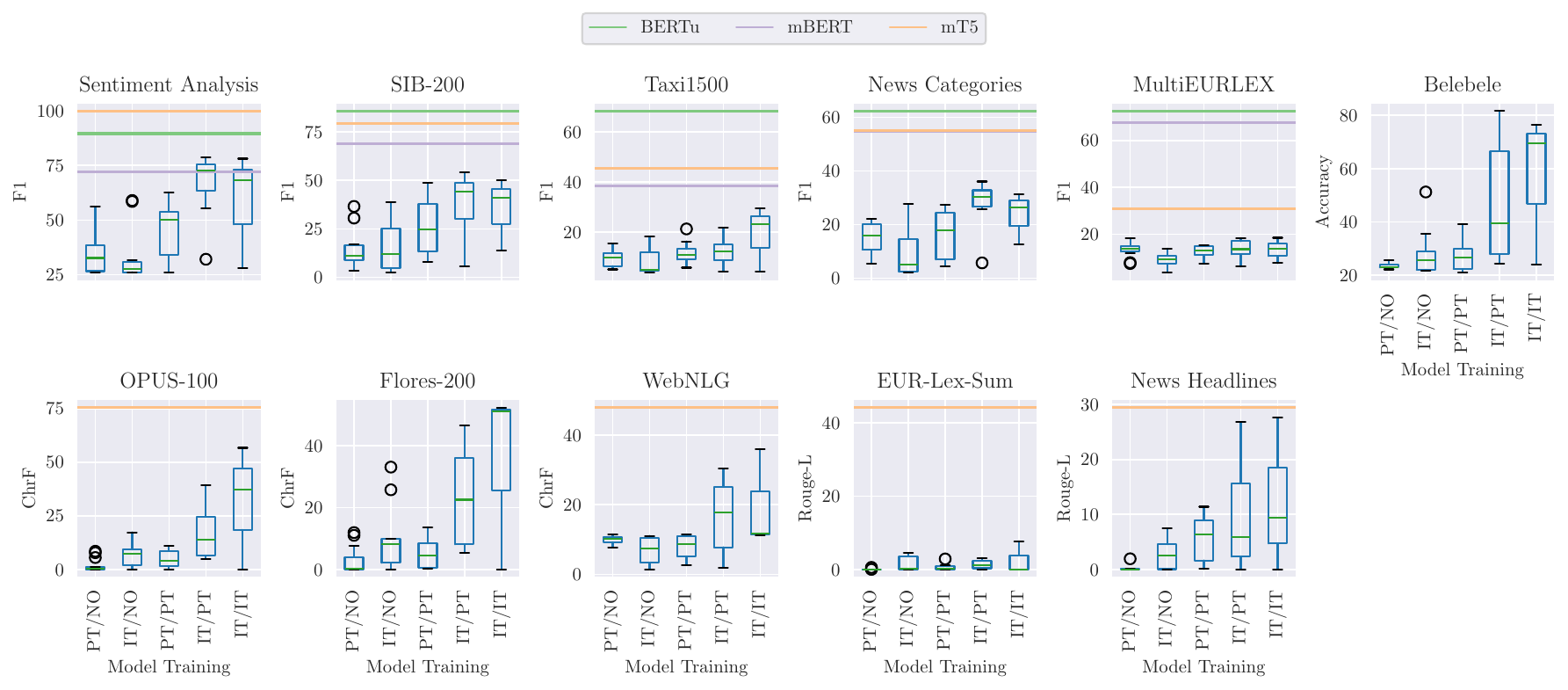}
    \caption{Zero-Shot}
    \label{figure:model_maltese_training_0shot_performance}
    \vspace{1em}
\end{subfigure}
\begin{subfigure}{\linewidth}
    \centering
    \includegraphics[scale=0.5]{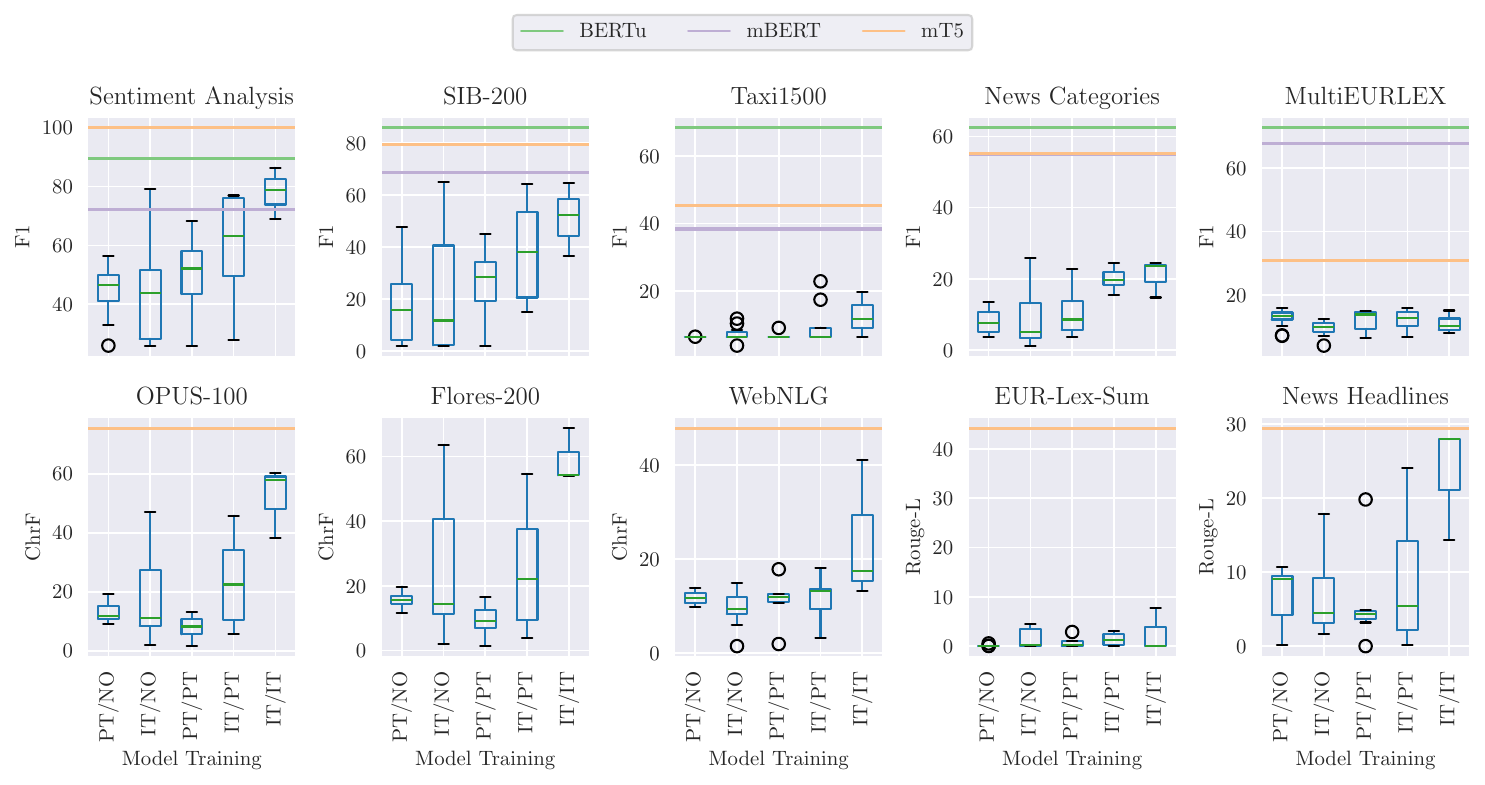}
    \caption{One-Shot}
    \label{figure:model_maltese_training_1shot_performance}
\end{subfigure}
\caption{Prompting performance distribution per task, with models grouped by different training types and Maltese training.}
\label{figure:model_maltese_training_performance}
\end{figure*}

We first examine the impact on model performance given its training on Maltese at different stages.
In the zero-shot experiments (Figure~\ref{figure:model_maltese_training_0shot_performance}), we observe that models exposed to Maltese during either \IT{} or \PT{} achieve better results, with the best overall results observed in the \IT{}/\IT{} category.
For \NLU{} tasks, \IT{}/\PT{} models perform on par with \IT{}/\IT{} models on News Categories and MultiEURLEX.
On \NLG{} tasks, \IT{}/\IT{} generally gives better results than \IT{}/\PT{}.

Similar to the observation made in Section~\ref{section:model_training}, \IT{} results in better overall scores than \PT{} models.
We also note that Maltese \PT{} is generally helpful, especially when comparing \PT{}/\PT{} against \PT{}/\NO{}.

For the one-shot experiments (Figure~\ref{figure:model_maltese_training_1shot_performance}), we note that \IT{}/\IT{} models outperform any other type of model more consistently and significantly, with the exception of the MultiEURLEX task.
\PT{}/\PT{} models are also generally less performant than \PT{}/\NO{} models.
This highlights that while instruction-tuning on a target language is beneficial, models sometimes need in-context examples to access their Maltese knowledge.

In Appendix~\ref{appendix:other_model_properties}, we also present further analyses exploring the relationship between performance and other dimensions such as a model's size and the number of languages on which it was trained.
We initially observed a slight correlation between performance and these variables.
However, model training on Maltese remains the main confounding variable, having a larger impact on performance.
When models trained on Maltese are excluded from these analyses, we observe that this correlation diminishes or is reversed.

\subsection{Maltese Prompts}
\label{section:maltese_prompts}

\begin{figure}[t!]
\begin{subfigure}{\linewidth}
    \centering
    \includegraphics[width=\linewidth]{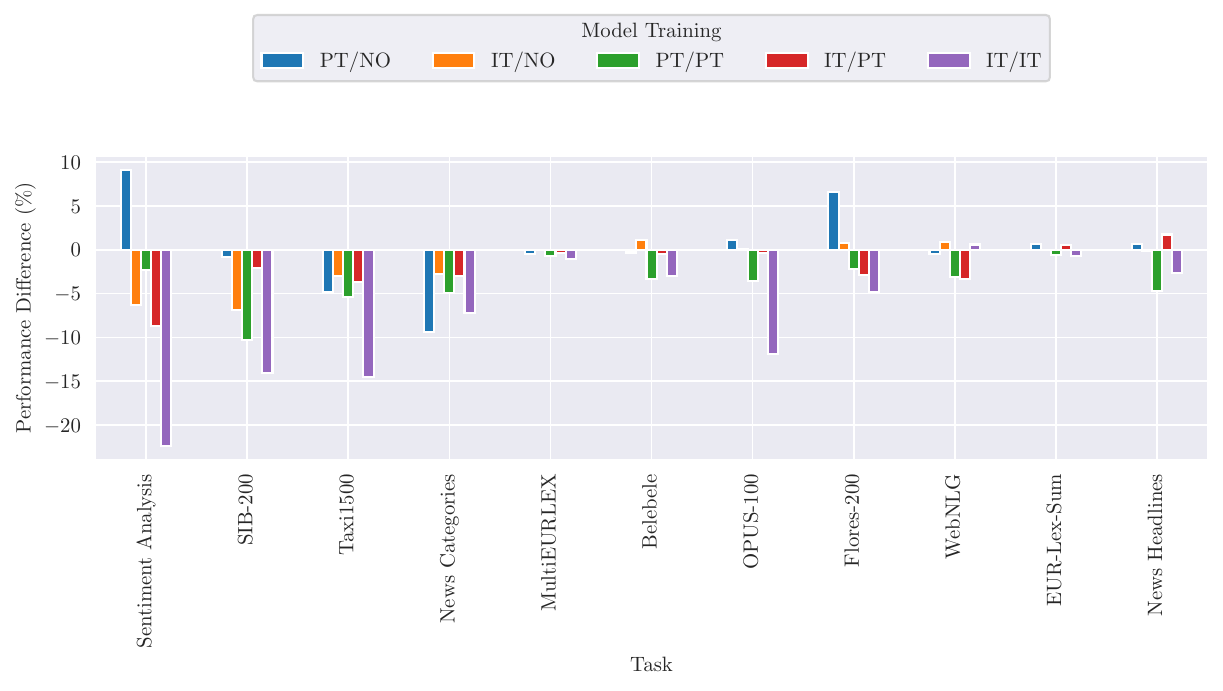}
    \caption{Zero-shot}
    \label{figure:prompt_language_0_shot_difference}
    \vspace{1em}
\end{subfigure}
\begin{subfigure}{\linewidth}
    \centering
    \includegraphics[width=\linewidth]{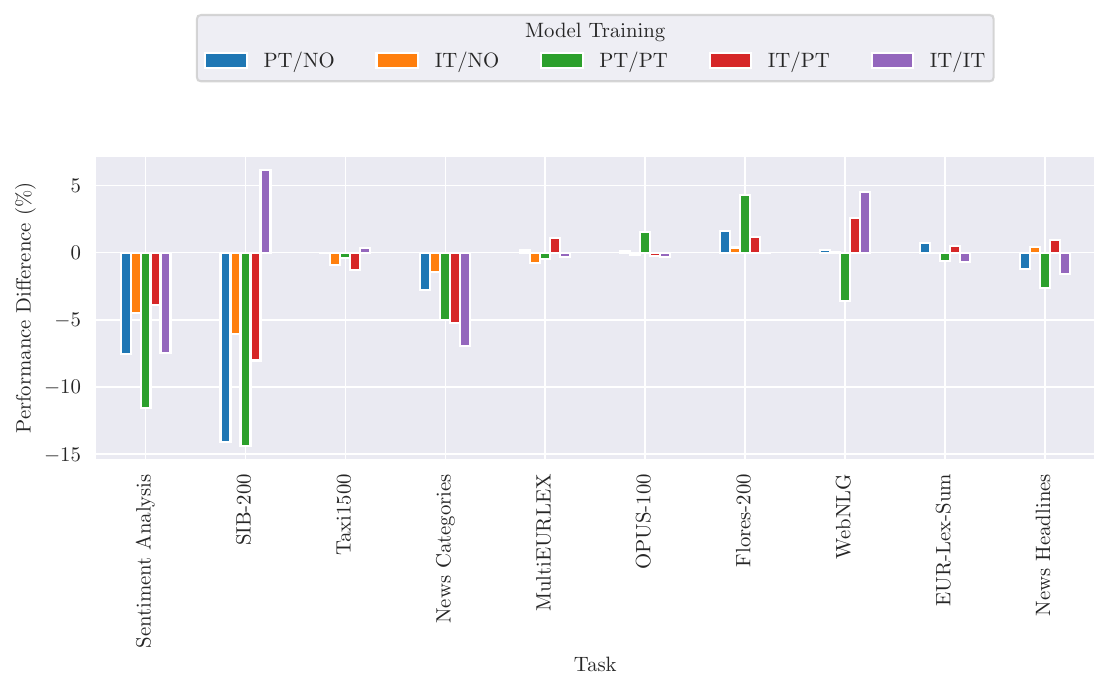}
    \caption{One-Shot}
    \label{figure:prompt_language_1shot_difference}
\end{subfigure}
\caption{Performance difference of prompting with English and Maltese instructions averaged across models with different training types and Maltese training.}
    \label{figure:prompt_language_difference}
\end{figure}

We now examine the impact of providing models with more Maltese text, not only in the form of in-context examples, but also by manually translating the instruction from English to Maltese.\footnote{
More detail regarding our prompt translation process is included in Appendix~\ref{appendix:prompts}.
}
Therefore, we repeat all previous prompting experiments using Maltese prompt templates.
Each score with English prompting is then subtracted from these new scores with Maltese prompting to get the performance difference, and the overall results are shown in Figure~\ref{figure:prompt_language_difference}.

We observe mixed results in both zero-shot and one-shot, but performance is generally worse with Maltese prompts.
However, models are negatively impacted in most \NLU{} tasks, regardless of the model's exposure to Maltese during its training.
With one-shot, the negative difference is even more pronounced, highlighting that in-context learning examples are better suited to prime the model to Maltese as opposed to language instructions.
\PT{}/\NO{} models seem to get significant improvements in Sentiment Analysis and Flores-200 with Maltese prompts in zero-shot, but this drastically diminishes in one-shot.

\IT{}/\IT{} models exhibit some of the worst degradations with Maltese prompts, particularly in zero-shot, even though these are the models that were exposed to Maltese the most.
However, although all models in this category -- EuroLLM Instruct and Aya-101 -- are exposed to Maltese examples during their \IT{}, the actual instructions are still in English.
This highlights the discrepancy between model performance and usability, since speaker populations of low-resource languages like Maltese would have to resort to providing English instructions in their interactions with these models.

\section{Efficiency}

The computational efficiency of fine-tuning and prompting is also an important factor to consider.
We select the following prompted models based on the best overall performance and the model architecture variety: 
Aya-101, mT0-XXL, and Llama 3 Instruct 8B.
We consider all fine-tuned models: BERTu, mBERT, and mT5-Small.

To estimate the computational requirements, we follow \citet{liu-etal-2022-tfew} and compute the Floating-Point Operations Per Second (FLOPs, \citealp{kaplan-etal-2020-scaling}) required for a single instance.
We take the median sequence length of an instance for each dataset and use it to calculate the inference FLOPs for a given model.
For fine-tuned models, we use only the raw input sequence in textual format.
For prompted models, we also include the accompanying instruction.
For the purpose of this analysis, we only consider zero-shot prompting with English instructions.
We also compute the training FLOPs for fine-tuned models by calculating the median sequence length on every training dataset, but also take into account the batch size and the total number of steps\footnote{This is calculated by averaging the total number of steps (including early stopping patience) across all runs.} during training.
We then take the average number of FLOPs across all tasks.
Since baseline models were not fine-tuned for Belebele and Flores-200, we do not include any calculations for these tasks in this analysis.
Table~\ref{table:flops} shows the resulting calculations.

\begin{table}[t!]
    \centering
    \footnotesize
    \setlength{\tabcolsep}{3pt} 
    \begin{tabular}{|l||r|r|}
        \hline
        \textbf{Model} & \textbf{Training FLOPs} & \textbf{Inference FLOPs} \\
        \hline\hline
        BERTu               & 1.54e16 & 3.28e10 \\
        mBERT               & 2.50e16 & 4.06e10 \\
        mT5-Small           & 7.19e15 & 7.14e09 \\
        \hline
        mT0-XXL             & 0       & 5.96e12 \\
        Aya-101             & 0       & 5.84e12 \\
        Llama 3 Instruct 8B & 0       & 5.06e13 \\
        \hline
    \end{tabular}
    \caption{
        Computational cost estimates in terms of Floating Point Operations (FLOPs).
    }
    \label{table:flops}
\end{table}

As expected, the inference cost of fine-tuned models is magnitudes smaller than that of prompted models due to the smaller model sizes.
We also note that for prompted models, as the number of few-shot examples increases, so does the inference cost.
The cost for fine-tuning models is also magnitudes larger than applying inference on a single instance.
However, as we apply inference on more examples, this initial upfront cost diminishes.

\begin{figure}[t!]
    \centering
    \includegraphics[width=\linewidth]{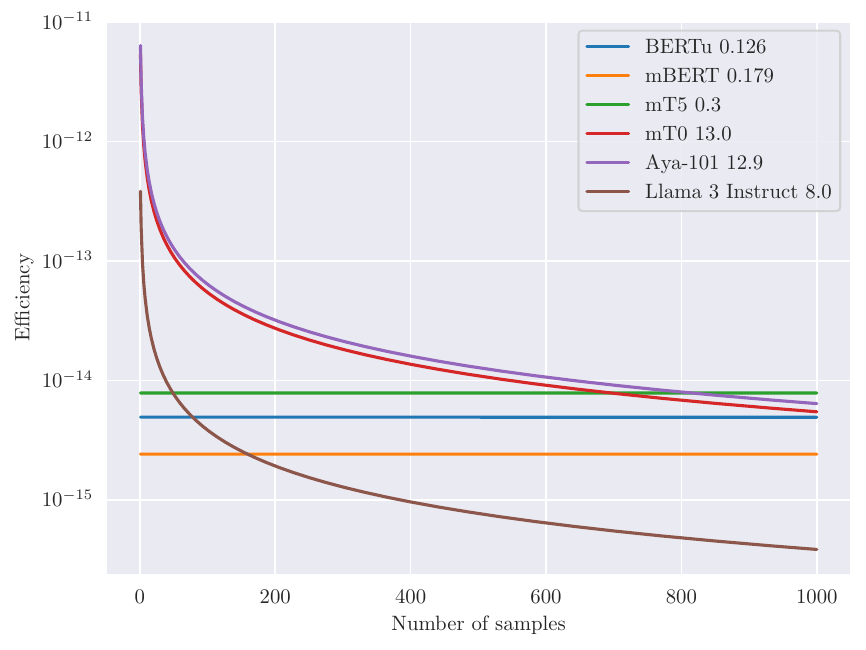}
    \caption{Model inference efficiency as a function of the performance and cost, as defined in Equation~\ref{equation:efficiency}.}
    \label{figure:efficiency}
\end{figure}

If we define efficiency as a function of the overall performance, the cost, and the number of inference samples as follows:
\begin{equation}
\label{equation:efficiency}
\frac{performance}{cost_{training} + cost_{inference} \times samples}
\end{equation}
then, as the number of samples increases, the efficiency of prompting larger models drastically decreases, as shown in Figure~\ref{figure:efficiency}.
Furthermore, the sheer size of the prompted models also necessitates more expensive hardware to store models on disk and load them into memory.

\section{Related Work}
\label{section:related_work}

Due to the lack of research on generative Maltese NLP, our research is primarily related to various multilingual benchmarking and evaluation works, although none of them include Maltese.
\citet{ahuja-etal-2023-mega} evaluate LLM performance on a newly developed benchmark covering 70 languages.
They find that low-resource languages are negatively affected by ill-fitted tokenisers and limited pre-training data used by the model.
\citet{zhang-etal-2024-impact} study the impact of few-shot prompting across 56 languages, finding that the few-shot performance does not always improve over zero-shot, depending on the model, task, and language.
\citet{asai-etal-2024-buffet} similarly analyse few-shot prompting performance but in a cross-lingual setup, finding that while few-shot generally helps, models perform particularly poorly on lower-resourced languages.
\citet{zhang-etal-2023-multilingual} analyse the performance on code-switched languages.
In their analysis, they find that both an increased number of shots and model size (when considering the same model family) generally improve performance.

Most of these works also fine-tune smaller BERT-based and/or mT5 models, similarly finding that they outperform prompted LLMs \cite{ahuja-etal-2023-mega, asai-etal-2024-buffet, zhang-etal-2023-multilingual}.
Likewise, \citet{lai-etal-2023-chatgpt} find that ChatGPT underperforms supervised SOTA models on most tasks.

Our research is complementary to these works as we look closer at the performance of a single language.
We also significantly scale the number of language models which allows us to study multiple factors affecting model performance.

\section{Conclusion}
\label{section:conclusion}

In this paper, we present a comprehensive evaluation of LLMs for Maltese, a low-resource language.
We introduce a new benchmark covering a total of 11 \NLU{} and \NLG{} tasks.
We carried out an evaluation of 55 publicly available models under zero-shot and one-shot prompting, revealing significant performance variations.
Our study highlights the need to improve the state of low-resource languages such as Maltese.

In analysing our results, we uncover several key trends.
Prompted models consistently lag behind smaller fine-tuned models, particularly in \NLG{} tasks where their performance is significantly lower.
Crucially, the level of exposure to Maltese in a model's training has the largest bearing on performance, especially in instruction-tuning.

Additionally, our efficiency analysis highlights the trade-offs between fine-tuning and prompting. While fine-tuning incurs higher initial costs, the inference cost is significantly lower.
The initial cost incurred to fine-tune smaller models quickly pays off with more inference instances, apart from the higher performance on downstream tasks.

These findings underscore the pressing need for more inclusive language models that better support low-resource languages at every stage of the training pipeline.
While LLMs offer strong generalisability, their limited performance on low-resource languages like Maltese reduces their usability in these scenarios.
For researchers with limited computational resources, fine-tuning smaller models presents a viable alternative to prompting larger models, despite the trade-off in generalisability.
Ultimately, our study calls for a more balanced approach to model development, ensuring that low-resource languages like Maltese receive adequate representation in the evolving landscape of LLMs.

\section{Limitations}
\label{section:limitations}

\paragraph{Models}
Although we consider a wide variety of models, the coverage of models trained on Maltese was very limited.
Moreover, due to the large number of models considered as well as computational constraints, we only choose models with no more than 15 billion parameters.
In addition, for our main analysis, we do not consider any commercial models, not only due to the prohibitive costs to conduct this evaluation, but also due to our constrained evaluation on \NLU{} tasks.
However, we present a limited comparative evaluation on ChatGPT in Appendix~\ref{appendix:chatgpt_result}, showing mixed results on the tasks tested compared to the fine-tuned baselines.

\paragraph{Datasets}
While our benchmark is certainly an improvement on the state of NLP for a low-resource language, certain issues impact our evaluation.
Firstly, our results are confounded by the amount of data in some tasks, which is why we strived to show per-task results as much as possible.
Despite the small training data sizes, we show that fine-tuned models still outperform the prompted large language models.
Secondly, the variance of tasks is also limited, as we have Topic Classification for the majority of our \NLU{} tasks, and Machine Translation dominates our \NLG{} tasks.
The latter is often a large source of instruction-tuning data for multilingual instruction-tuning, and often the only data considered when Maltese was used for instruction-tuning \cite{ustun-etal-2024-aya}.
Thirdly, the domains of these datasets are quite narrow as they are mostly composed of data derived from news articles, EU legislative documents, and Wikimedia.
All in all, we hope that our work raises awareness on the importance of developing newer and more diverse datasets for low-resource languages.

\paragraph{Prompt Engineering}
Various works have shown that different models can be optimised with different prompts (\citealp{zhao-etal-2021-calibrate, jiang-etal-2020-prompt-engineering, shin-etal-2020-autoprompt}; inter alia).
We sidestep this by mostly using prompt templates from previous works (see Appendix~\ref{appendix:prompts} for more details).
We highlight that we did consider prompting in Maltese (Section~\ref{section:maltese_prompts}), which is often understudied in the context of multilingual evaluation \cite{ahuja-etal-2023-mega, asai-etal-2024-buffet}.
We did not experiment with a larger number of shots to keep our experiments sustainable.

\paragraph{Unconsidered Model Properties}
We tried to analyse many possible dimensions but still had to limit our search space.
Despite looking at models exposed to Maltese training, we largely treated this as a categorical variable.
However, the raw amount of tokens, as well as the proportion in relation to the rest of the training data, would have a large bearing on performance.
In a similar vein, we did not consider the different scales and quality of the datasets used for training different models.
We also do not factor for different kinds of training processes used during instruction-tuning such as Reinforcement Learning with Human Feedback \cite{ziegler-etal-2020-rlhf} and Direct Preference Optimisation \cite{rafailov-etal-2023-dpo}.
As we make this data available, we encourage future work to analyse different dimensions not considered in this work.

\paragraph{Dataset Contamination}
It is likely that models have been exposed to language data that is not included in the figures listed in Table~\ref{table:models} \cite{blevins-zettlemoyer-2022-language, muennighoff-etal-2023-crosslingual}, but we do not account for it.
Since our benchmarks are based on publicly available datasets, it is likely that increased performance in some models is due to data contamination during training.
While this can be accidental, instruction-tuned models may have used certain datasets deliberately.
Hence, higher scores would be attributed to the model's training on that task, rather than its capabilities to generalise to unseen tasks.
Dataset contamination is also an active area of research \cite{blevins-zettlemoyer-2022-language, balloccu-etal-2024-leak}, and we therefore treat models as black-box systems in this regard.

\section{Ethics Statement}
\label{section:ethics}

We inherit any biases that may be present in the data and language models that we use.
The new generative models fine-tuned on Maltese data could be used to produce text that inherit these biases.
However, given their relatively low performance on \NLG{} tasks and the fact that we train these using publicly available resources, we do not foresee any major risks.

\section*{Acknowledgements}

We acknowledge support from the LT-Bridge Project (GA 952194) and DFKI for access to the Virtual Laboratory.
We further acknowledge funding by Malta Enterprise Research and Development Scheme.

\bibliography{anthology,custom}

\begin{thebibliography}{51}
\providecommand{\natexlab}[1]{#1}

\bibitem[{Abela et~al.(2024)Abela, Micallef, Tanti, and Borg}]{abela-etal-2024-tokenisation}
Kurt Abela, Kurt Micallef, Marc Tanti, and Claudia Borg. 2024.
\newblock \href {https://doi.org/10.18653/v1/2024.loresmt-1.11} {Tokenisation in machine translation does matter: The impact of different tokenisation approaches for {M}altese}.
\newblock In \emph{Proceedings of the Seventh Workshop on Technologies for Machine Translation of Low-Resource Languages (LoResMT 2024)}, pages 109--120, Bangkok, Thailand. Association for Computational Linguistics.

\bibitem[{Adelani et~al.(2024)Adelani, Liu, Shen, Vassilyev, Alabi, Mao, Gao, and Lee}]{adelani-etal-2024-sib}
David~Ifeoluwa Adelani, Hannah Liu, Xiaoyu Shen, Nikita Vassilyev, Jesujoba~O. Alabi, Yanke Mao, Haonan Gao, and En-Shiun~Annie Lee. 2024.
\newblock \href {https://aclanthology.org/2024.eacl-long.14/} {{SIB}-200: A simple, inclusive, and big evaluation dataset for topic classification in 200+ languages and dialects}.
\newblock In \emph{Proceedings of the 18th Conference of the European Chapter of the Association for Computational Linguistics (Volume 1: Long Papers)}, pages 226--245, St. Julian{'}s, Malta. Association for Computational Linguistics.

\bibitem[{Ahuja et~al.(2023)Ahuja, Diddee, Hada, Ochieng, Ramesh, Jain, Nambi, Ganu, Segal, Ahmed, Bali, and Sitaram}]{ahuja-etal-2023-mega}
Kabir Ahuja, Harshita Diddee, Rishav Hada, Millicent Ochieng, Krithika Ramesh, Prachi Jain, Akshay Nambi, Tanuja Ganu, Sameer Segal, Mohamed Ahmed, et~al. 2023.
\newblock \href {https://doi.org/10.18653/v1/2023.emnlp-main.258} {{MEGA}: Multilingual evaluation of generative {AI}}.
\newblock In \emph{Proceedings of the 2023 Conference on Empirical Methods in Natural Language Processing}, pages 4232--4267, Singapore. Association for Computational Linguistics.

\bibitem[{Ali et~al.(2024)Ali, Fromm, Thellmann, Ebert, Weber, Rutmann, Jain, Lübbering, Steinigen, Leveling, Klug, Buschhoff, Jurkschat, Abdelwahab, Stein, Sylla, Denisov, Brandizzi, Saleem, Bhowmick, Helmer, John, Suarez, Ostendorff, Jude, Manjunath, Weinbach, Penke, Filatov, Asaadi, Barth, Sifa, Küch, Herten, Jäkel, Rehm, Kesselheim, Köhler, and Flores-Herr}]{ali-etal-2024-teuken}
Mehdi Ali, Michael Fromm, Klaudia Thellmann, Jan Ebert, Alexander~Arno Weber, Richard Rutmann, Charvi Jain, Max Lübbering, Daniel Steinigen, Johannes Leveling, et~al. 2024.
\newblock \href {https://arxiv.org/abs/2410.03730} {{Teuken-7B-Base} \& {Teuken-7B-Instruct}: Towards {E}uropean {LLMs}}.
\newblock \emph{Preprint}, arXiv:2410.03730.

\bibitem[{Aryabumi et~al.(2024)Aryabumi, Dang, Talupuru, Dash, Cairuz, Lin, Venkitesh, Smith, Campos, Tan, Marchisio, Bartolo, Ruder, Locatelli, Kreutzer, Frosst, Gomez, Blunsom, Fadaee, Üstün, and Hooker}]{aryabumi-etal-2024-aya23}
Viraat Aryabumi, John Dang, Dwarak Talupuru, Saurabh Dash, David Cairuz, Hangyu Lin, Bharat Venkitesh, Madeline Smith, Jon~Ander Campos, Yi~Chern Tan, et~al. 2024.
\newblock \href {https://arxiv.org/abs/2405.15032} {{Aya 23}: Open weight releases to further multilingual progress}.
\newblock \emph{Preprint}, arXiv:2405.15032.

\bibitem[{Asai et~al.(2024)Asai, Kudugunta, Yu, Blevins, Gonen, Reid, Tsvetkov, Ruder, and Hajishirzi}]{asai-etal-2024-buffet}
Akari Asai, Sneha Kudugunta, Xinyan Yu, Terra Blevins, Hila Gonen, Machel Reid, Yulia Tsvetkov, Sebastian Ruder, and Hannaneh Hajishirzi. 2024.
\newblock \href {https://doi.org/10.18653/v1/2024.naacl-long.100} {{BUFFET}: Benchmarking large language models for few-shot cross-lingual transfer}.
\newblock In \emph{Proceedings of the 2024 Conference of the North American Chapter of the Association for Computational Linguistics: Human Language Technologies (Volume 1: Long Papers)}, pages 1771--1800, Mexico City, Mexico. Association for Computational Linguistics.

\bibitem[{Aumiller et~al.(2022)Aumiller, Chouhan, and Gertz}]{aumiller-etal-2022-eur}
Dennis Aumiller, Ashish Chouhan, and Michael Gertz. 2022.
\newblock \href {https://doi.org/10.18653/v1/2022.emnlp-main.519} {{EUR}-lex-sum: A multi- and cross-lingual dataset for long-form summarization in the legal domain}.
\newblock In \emph{Proceedings of the 2022 Conference on Empirical Methods in Natural Language Processing}, pages 7626--7639, Abu Dhabi, United Arab Emirates. Association for Computational Linguistics.

\bibitem[{Balloccu et~al.(2024)Balloccu, Schmidtov{\'a}, Lango, and Dusek}]{balloccu-etal-2024-leak}
Simone Balloccu, Patr{\'i}cia Schmidtov{\'a}, Mateusz Lango, and Ondrej Dusek. 2024.
\newblock \href {https://aclanthology.org/2024.eacl-long.5/} {Leak, cheat, repeat: Data contamination and evaluation malpractices in closed-source {LLM}s}.
\newblock In \emph{Proceedings of the 18th Conference of the European Chapter of the Association for Computational Linguistics (Volume 1: Long Papers)}, pages 67--93, St. Julian{'}s, Malta. Association for Computational Linguistics.

\bibitem[{Bandarkar et~al.(2024)Bandarkar, Liang, Muller, Artetxe, Shukla, Husa, Goyal, Krishnan, Zettlemoyer, and Khabsa}]{bandarkar-etal-2024-belebele}
Lucas Bandarkar, Davis Liang, Benjamin Muller, Mikel Artetxe, Satya~Narayan Shukla, Donald Husa, Naman Goyal, Abhinandan Krishnan, Luke Zettlemoyer, and Madian Khabsa. 2024.
\newblock \href {https://doi.org/10.18653/v1/2024.acl-long.44} {The belebele benchmark: a parallel reading comprehension dataset in 122 language variants}.
\newblock In \emph{Proceedings of the 62nd Annual Meeting of the Association for Computational Linguistics (Volume 1: Long Papers)}, pages 749--775, Bangkok, Thailand. Association for Computational Linguistics.

\bibitem[{{BigScience Workshop} et~al.(2023){BigScience Workshop}, Scao, Fan, Akiki, Pavlick, Ilić, Hesslow, Castagné, Luccioni, Yvon, Gallé, Tow, Rush, Biderman, Webson, Ammanamanchi, Wang, Sagot, Muennighoff, del Moral, Ruwase, Bawden, Bekman, McMillan-Major, Beltagy, Nguyen, Saulnier, Tan, Suarez, Sanh, Laurençon, Jernite, Launay, Mitchell, Raffel, Gokaslan, Simhi, Soroa, Aji, Alfassy, Rogers, Nitzav, Xu, Mou, Emezue, Klamm, Leong, van Strien, Adelani, Radev, Ponferrada, Levkovizh, Kim, Natan, Toni, Dupont, Kruszewski, Pistilli, Elsahar, Benyamina, Tran, Yu, Abdulmumin, Johnson, Gonzalez-Dios, de~la Rosa, Chim, Dodge, Zhu, Chang, Frohberg, Tobing, Bhattacharjee, Almubarak, Chen, Lo, Werra, Weber, Phan, allal, Tanguy, Dey, Muñoz, Masoud, Grandury, Šaško, Huang, Coavoux, Singh, Jiang, Vu, Jauhar, Ghaleb, Subramani, Kassner, Khamis, Nguyen, Espejel, de~Gibert, Villegas, Henderson, Colombo, Amuok, Lhoest, Harliman, Bommasani, López, Ribeiro, Osei, Pyysalo, Nagel, Bose, Muhammad, Sharma, Longpre,
  Nikpoor, Silberberg, Pai, Zink, Torrent, Schick, Thrush, Danchev, Nikoulina, Laippala, Lepercq, Prabhu, Alyafeai, Talat, Raja, Heinzerling, Si, Taşar, Salesky, Mielke, Lee, Sharma, Santilli, Chaffin, Stiegler, Datta, Szczechla, Chhablani, Wang, Pandey, Strobelt, Fries, Rozen, Gao, Sutawika, Bari, Al-shaibani, Manica, Nayak, Teehan, Albanie, Shen, Ben-David, Bach, Kim, Bers, Fevry, Neeraj, Thakker, Raunak, Tang, Yong, Sun, Brody, Uri, Tojarieh, Roberts, Chung, Tae, Phang, Press, Li, Narayanan, Bourfoune, Casper, Rasley, Ryabinin, Mishra, Zhang, Shoeybi, Peyrounette, Patry, Tazi, Sanseviero, von Platen, Cornette, Lavallée, Lacroix, Rajbhandari, Gandhi, Smith, Requena, Patil, Dettmers, Baruwa, Singh, Cheveleva, Ligozat, Subramonian, Névéol, Lovering, Garrette, Tunuguntla, Reiter, Taktasheva, Voloshina, Bogdanov, Winata, Schoelkopf, Kalo, Novikova, Forde, Clive, Kasai, Kawamura, Hazan, Carpuat, Clinciu, Kim, Cheng, Serikov, Antverg, van~der Wal, Zhang, Zhang, Gehrmann, Mirkin, Pais, Shavrina, Scialom, Yun,
  Limisiewicz, Rieser, Protasov, Mikhailov, Pruksachatkun, Belinkov, Bamberger, Kasner, Rueda, Pestana, Feizpour, Khan, Faranak, Santos, Hevia, Unldreaj, Aghagol, Abdollahi, Tammour, HajiHosseini, Behroozi, Ajibade, Saxena, Ferrandis, McDuff, Contractor, Lansky, David, Kiela, Nguyen, Tan, Baylor, Ozoani, Mirza, Ononiwu, Rezanejad, Jones, Bhattacharya, Solaiman, Sedenko, Nejadgholi, Passmore, Seltzer, Sanz, Dutra, Samagaio, Elbadri, Mieskes, Gerchick, Akinlolu, McKenna, Qiu, Ghauri, Burynok, Abrar, Rajani, Elkott, Fahmy, Samuel, An, Kromann, Hao, Alizadeh, Shubber, Wang, Roy, Viguier, Le, Oyebade, Le, Yang, Nguyen, Kashyap, Palasciano, Callahan, Shukla, Miranda-Escalada, Singh, Beilharz, Wang, Brito, Zhou, Jain, Xu, Fourrier, Periñán, Molano, Yu, Manjavacas, Barth, Fuhrimann, Altay, Bayrak, Burns, Vrabec, Bello, Dash, Kang, Giorgi, Golde, Posada, Sivaraman, Bulchandani, Liu, Shinzato, de~Bykhovetz, Takeuchi, Pàmies, Castillo, Nezhurina, Sänger, Samwald, Cullan, Weinberg, Wolf, Mihaljcic, Liu, Freidank,
  Kang, Seelam, Dahlberg, Broad, Muellner, Fung, Haller, Chandrasekhar, Eisenberg, Martin, Canalli, Su, Su, Cahyawijaya, Garda, Deshmukh, Mishra, Kiblawi, Ott, Sang-aroonsiri, Kumar, Schweter, Bharati, Laud, Gigant, Kainuma, Kusa, Labrak, Bajaj, Venkatraman, Xu, Xu, Xu, Tan, Xie, Ye, Bras, Belkada, and Wolf}]{bigscience-2023-bloom}
{BigScience Workshop}, Teven~Le Scao, Angela Fan, Christopher Akiki, Ellie Pavlick, Suzana Ilić, Daniel Hesslow, Roman Castagné, Alexandra~Sasha Luccioni, François Yvon, et~al. 2023.
\newblock \href {https://arxiv.org/abs/2211.05100} {{BLOOM}: A 176{B}-parameter open-access multilingual language model}.
\newblock \emph{Preprint}, arXiv:2211.05100.

\bibitem[{Blevins and Zettlemoyer(2022)}]{blevins-zettlemoyer-2022-language}
Terra Blevins and Luke Zettlemoyer. 2022.
\newblock \href {https://doi.org/10.18653/v1/2022.emnlp-main.233} {Language contamination helps explains the cross-lingual capabilities of {E}nglish pretrained models}.
\newblock In \emph{Proceedings of the 2022 Conference on Empirical Methods in Natural Language Processing}, pages 3563--3574, Abu Dhabi, United Arab Emirates. Association for Computational Linguistics.

\bibitem[{Brown et~al.(2020)Brown, Mann, Ryder, Subbiah, Kaplan, Dhariwal, Neelakantan, Shyam, Sastry, Askell, Agarwal, Herbert-Voss, Krueger, Henighan, Child, Ramesh, Ziegler, Wu, Winter, Hesse, Chen, Sigler, Litwin, Gray, Chess, Clark, Berner, McCandlish, Radford, Sutskever, and Amodei}]{brown-etal-2020-gpt3}
Tom Brown, Benjamin Mann, Nick Ryder, Melanie Subbiah, Jared~D Kaplan, Prafulla Dhariwal, Arvind Neelakantan, Pranav Shyam, Girish Sastry, Amanda Askell, et~al. 2020.
\newblock \href {https://proceedings.neurips.cc/paper_files/paper/2020/file/1457c0d6bfcb4967418bfb8ac142f64a-Paper.pdf} {Language models are few-shot learners}.
\newblock In \emph{Advances in Neural Information Processing Systems}, volume~33, pages 1877--1901. Curran Associates, Inc.

\bibitem[{Chalkidis et~al.(2021)Chalkidis, Fergadiotis, and Androutsopoulos}]{chalkidis-etal-2021-multieurlex}
Ilias Chalkidis, Manos Fergadiotis, and Ion Androutsopoulos. 2021.
\newblock \href {https://doi.org/10.18653/v1/2021.emnlp-main.559} {{M}ulti{EURLEX} - a multi-lingual and multi-label legal document classification dataset for zero-shot cross-lingual transfer}.
\newblock In \emph{Proceedings of the 2021 Conference on Empirical Methods in Natural Language Processing}, pages 6974--6996, Online and Punta Cana, Dominican Republic. Association for Computational Linguistics.

\bibitem[{Chau et~al.(2020)Chau, Lin, and Smith}]{chau-etal-2020-parsing}
Ethan~C. Chau, Lucy~H. Lin, and Noah~A. Smith. 2020.
\newblock \href {https://doi.org/10.18653/v1/2020.findings-emnlp.118} {Parsing with multilingual {BERT}, a small corpus, and a small treebank}.
\newblock In \emph{Findings of the Association for Computational Linguistics: EMNLP 2020}, pages 1324--1334, Online. Association for Computational Linguistics.

\bibitem[{Chaudhary et~al.(2024)Chaudhary, Micallef, and Borg}]{chaudhary-etal-2024-topic}
Amit~Kumar Chaudhary, Kurt Micallef, and Claudia Borg. 2024.
\newblock \href {https://aclanthology.org/2024.lrec-main.1414/} {Topic classification and headline generation for {M}altese using a public news corpus}.
\newblock In \emph{Proceedings of the 2024 Joint International Conference on Computational Linguistics, Language Resources and Evaluation (LREC-COLING 2024)}, pages 16274--16281, Torino, Italia. ELRA and ICCL.

\bibitem[{Cripwell et~al.(2023)Cripwell, Belz, Gardent, Gatt, Borg, Borg, Judge, Lorandi, Nikiforovskaya, and Soto~Martinez}]{cripwell-etal-2023-2023}
Liam Cripwell, Anya Belz, Claire Gardent, Albert Gatt, Claudia Borg, Marthese Borg, John Judge, Michela Lorandi, Anna Nikiforovskaya, and William Soto~Martinez. 2023.
\newblock \href {https://aclanthology.org/2023.mmnlg-1.6/} {The 2023 {W}eb{NLG} shared task on low resource languages. overview and evaluation results ({W}eb{NLG} 2023)}.
\newblock In \emph{Proceedings of the Workshop on Multimodal, Multilingual Natural Language Generation and Multilingual WebNLG Challenge (MM-NLG 2023)}, pages 55--66, Prague, Czech Republic. Association for Computational Linguistics.

\bibitem[{Devlin et~al.(2019)Devlin, Chang, Lee, and Toutanova}]{devlin-etal-2019-bert}
Jacob Devlin, Ming-Wei Chang, Kenton Lee, and Kristina Toutanova. 2019.
\newblock \href {https://doi.org/10.18653/v1/N19-1423} {{BERT}: Pre-training of deep bidirectional transformers for language understanding}.
\newblock In \emph{Proceedings of the 2019 Conference of the North {A}merican Chapter of the Association for Computational Linguistics: Human Language Technologies, Volume 1 (Long and Short Papers)}, pages 4171--4186, Minneapolis, Minnesota. Association for Computational Linguistics.

\bibitem[{Gao et~al.(2024)Gao, Tow, Abbasi, Biderman, Black, DiPofi, Foster, Golding, Hsu, Le~Noac'h, Li, McDonell, Muennighoff, Ociepa, Phang, Reynolds, Schoelkopf, Skowron, Sutawika, Tang, Thite, Wang, Wang, and Zou}]{lm-evaluation-harness}
Leo Gao, Jonathan Tow, Baber Abbasi, Stella Biderman, Sid Black, Anthony DiPofi, Charles Foster, Laurence Golding, Jeffrey Hsu, Alain Le~Noac'h, et~al. 2024.
\newblock \href {https://doi.org/10.5281/zenodo.12608602} {A framework for few-shot language model evaluation}.

\bibitem[{{Gemma Team} et~al.(2024){Gemma Team}, Riviere, Pathak, Sessa, Hardin, Bhupatiraju, Hussenot, Mesnard, Shahriari, Ramé, Ferret, Liu, Tafti, Friesen, Casbon, Ramos, Kumar, Lan, Jerome, Tsitsulin, Vieillard, Stanczyk, Girgin, Momchev, Hoffman, Thakoor, Grill, Neyshabur, Bachem, Walton, Severyn, Parrish, Ahmad, Hutchison, Abdagic, Carl, Shen, Brock, Coenen, Laforge, Paterson, Bastian, Piot, Wu, Royal, Chen, Kumar, Perry, Welty, Choquette-Choo, Sinopalnikov, Weinberger, Vijaykumar, Rogozińska, Herbison, Bandy, Wang, Noland, Moreira, Senter, Eltyshev, Visin, Rasskin, Wei, Cameron, Martins, Hashemi, Klimczak-Plucińska, Batra, Dhand, Nardini, Mein, Zhou, Svensson, Stanway, Chan, Zhou, Carrasqueira, Iljazi, Becker, Fernandez, van Amersfoort, Gordon, Lipschultz, Newlan, yeong Ji, Mohamed, Badola, Black, Millican, McDonell, Nguyen, Sodhia, Greene, Sjoesund, Usui, Sifre, Heuermann, Lago, McNealus, Soares, Kilpatrick, Dixon, Martins, Reid, Singh, Iverson, Görner, Velloso, Wirth, Davidow, Miller, Rahtz,
  Watson, Risdal, Kazemi, Moynihan, Zhang, Kahng, Park, Rahman, Khatwani, Dao, Bardoliwalla, Devanathan, Dumai, Chauhan, Wahltinez, Botarda, Barnes, Barham, Michel, Jin, Georgiev, Culliton, Kuppala, Comanescu, Merhej, Jana, Rokni, Agarwal, Mullins, Saadat, Carthy, Cogan, Perrin, Arnold, Krause, Dai, Garg, Sheth, Ronstrom, Chan, Jordan, Yu, Eccles, Hennigan, Kocisky, Doshi, Jain, Yadav, Meshram, Dharmadhikari, Barkley, Wei, Ye, Han, Kwon, Xu, Shen, Gong, Wei, Cotruta, Kirk, Rao, Giang, Peran, Warkentin, Collins, Barral, Ghahramani, Hadsell, Sculley, Banks, Dragan, Petrov, Vinyals, Dean, Hassabis, Kavukcuoglu, Farabet, Buchatskaya, Borgeaud, Fiedel, Joulin, Kenealy, Dadashi, and Andreev}]{gemma-2024-gemma2}
{Gemma Team}, Morgane Riviere, Shreya Pathak, Pier~Giuseppe Sessa, Cassidy Hardin, Surya Bhupatiraju, Léonard Hussenot, Thomas Mesnard, Bobak Shahriari, Alexandre Ramé, et~al. 2024.
\newblock \href {https://arxiv.org/abs/2408.00118} {{Gemma 2}: Improving open language models at a practical size}.
\newblock \emph{Preprint}, arXiv:2408.00118.

\bibitem[{Gonzalez-Agirre et~al.(2025)Gonzalez-Agirre, Pàmies, Llop, Baucells, Dalt, Tamayo, Saiz, Espuña, Prats, Aula-Blasco, Mina, Pikabea, Rubio, Shvets, Sallés, Lacunza, Palomar, Falcão, Tormo, Vasquez-Reina, Marimon, Pareras, Ruiz-Fernández, and Villegas}]{gonzalez-agirre-etal-2025-salamandra}
Aitor Gonzalez-Agirre, Marc Pàmies, Joan Llop, Irene Baucells, Severino~Da Dalt, Daniel Tamayo, José~Javier Saiz, Ferran Espuña, Jaume Prats, Javier Aula-Blasco, et~al. 2025.
\newblock \href {https://arxiv.org/abs/2502.08489} {Salamandra technical report}.
\newblock \emph{Preprint}, arXiv:2502.08489.

\bibitem[{Grattafiori et~al.(2024)Grattafiori, Dubey, Jauhri, Pandey, Kadian, Al-Dahle, Letman, Mathur, Schelten, Vaughan, Yang, Fan, Goyal, Hartshorn, Yang, Mitra, Sravankumar, Korenev, Hinsvark, Rao, Zhang, Rodriguez, Gregerson, Spataru, Roziere, Biron, Tang, Chern, Caucheteux, Nayak, Bi, Marra, McConnell, Keller, Touret, Wu, Wong, Ferrer, Nikolaidis, Allonsius, Song, Pintz, Livshits, Wyatt, Esiobu, Choudhary, Mahajan, Garcia-Olano, Perino, Hupkes, Lakomkin, AlBadawy, Lobanova, Dinan, Smith, Radenovic, Guzmán, Zhang, Synnaeve, Lee, Anderson, Thattai, Nail, Mialon, Pang, Cucurell, Nguyen, Korevaar, Xu, Touvron, Zarov, Ibarra, Kloumann, Misra, Evtimov, Zhang, Copet, Lee, Geffert, Vranes, Park, Mahadeokar, Shah, van~der Linde, Billock, Hong, Lee, Fu, Chi, Huang, Liu, Wang, Yu, Bitton, Spisak, Park, Rocca, Johnstun, Saxe, Jia, Alwala, Prasad, Upasani, Plawiak, Li, Heafield, Stone, El-Arini, Iyer, Malik, Chiu, Bhalla, Lakhotia, Rantala-Yeary, van~der Maaten, Chen, Tan, Jenkins, Martin, Madaan, Malo, Blecher,
  Landzaat, de~Oliveira, Muzzi, Pasupuleti, Singh, Paluri, Kardas, Tsimpoukelli, Oldham, Rita, Pavlova, Kambadur, Lewis, Si, Singh, Hassan, Goyal, Torabi, Bashlykov, Bogoychev, Chatterji, Zhang, Duchenne, Çelebi, Alrassy, Zhang, Li, Vasic, Weng, Bhargava, Dubal, Krishnan, Koura, Xu, He, Dong, Srinivasan, Ganapathy, Calderer, Cabral, Stojnic, Raileanu, Maheswari, Girdhar, Patel, Sauvestre, Polidoro, Sumbaly, Taylor, Silva, Hou, Wang, Hosseini, Chennabasappa, Singh, Bell, Kim, Edunov, Nie, Narang, Raparthy, Shen, Wan, Bhosale, Zhang, Vandenhende, Batra, Whitman, Sootla, Collot, Gururangan, Borodinsky, Herman, Fowler, Sheasha, Georgiou, Scialom, Speckbacher, Mihaylov, Xiao, Karn, Goswami, Gupta, Ramanathan, Kerkez, Gonguet, Do, Vogeti, Albiero, Petrovic, Chu, Xiong, Fu, Meers, Martinet, Wang, Wang, Tan, Xia, Xie, Jia, Wang, Goldschlag, Gaur, Babaei, Wen, Song, Zhang, Li, Mao, Coudert, Yan, Chen, Papakipos, Singh, Srivastava, Jain, Kelsey, Shajnfeld, Gangidi, Victoria, Goldstand, Menon, Sharma, Boesenberg,
  Baevski, Feinstein, Kallet, Sangani, Teo, Yunus, Lupu, Alvarado, Caples, Gu, Ho, Poulton, Ryan, Ramchandani, Dong, Franco, Goyal, Saraf, Chowdhury, Gabriel, Bharambe, Eisenman, Yazdan, James, Maurer, Leonhardi, Huang, Loyd, Paola, Paranjape, Liu, Wu, Ni, Hancock, Wasti, Spence, Stojkovic, Gamido, Montalvo, Parker, Burton, Mejia, Liu, Wang, Kim, Zhou, Hu, Chu, Cai, Tindal, Feichtenhofer, Gao, Civin, Beaty, Kreymer, Li, Adkins, Xu, Testuggine, David, Parikh, Liskovich, Foss, Wang, Le, Holland, Dowling, Jamil, Montgomery, Presani, Hahn, Wood, Le, Brinkman, Arcaute, Dunbar, Smothers, Sun, Kreuk, Tian, Kokkinos, Ozgenel, Caggioni, Kanayet, Seide, Florez, Schwarz, Badeer, Swee, Halpern, Herman, Sizov, Guangyi, Zhang, Lakshminarayanan, Inan, Shojanazeri, Zou, Wang, Zha, Habeeb, Rudolph, Suk, Aspegren, Goldman, Zhan, Damlaj, Molybog, Tufanov, Leontiadis, Veliche, Gat, Weissman, Geboski, Kohli, Lam, Asher, Gaya, Marcus, Tang, Chan, Zhen, Reizenstein, Teboul, Zhong, Jin, Yang, Cummings, Carvill, Shepard, McPhie,
  Torres, Ginsburg, Wang, Wu, U, Saxena, Khandelwal, Zand, Matosich, Veeraraghavan, Michelena, Li, Jagadeesh, Huang, Chawla, Huang, Chen, Garg, A, Silva, Bell, Zhang, Guo, Yu, Moshkovich, Wehrstedt, Khabsa, Avalani, Bhatt, Mankus, Hasson, Lennie, Reso, Groshev, Naumov, Lathi, Keneally, Liu, Seltzer, Valko, Restrepo, Patel, Vyatskov, Samvelyan, Clark, Macey, Wang, Hermoso, Metanat, Rastegari, Bansal, Santhanam, Parks, White, Bawa, Singhal, Egebo, Usunier, Mehta, Laptev, Dong, Cheng, Chernoguz, Hart, Salpekar, Kalinli, Kent, Parekh, Saab, Balaji, Rittner, Bontrager, Roux, Dollar, Zvyagina, Ratanchandani, Yuvraj, Liang, Alao, Rodriguez, Ayub, Murthy, Nayani, Mitra, Parthasarathy, Li, Hogan, Battey, Wang, Howes, Rinott, Mehta, Siby, Bondu, Datta, Chugh, Hunt, Dhillon, Sidorov, Pan, Mahajan, Verma, Yamamoto, Ramaswamy, Lindsay, Lindsay, Feng, Lin, Zha, Patil, Shankar, Zhang, Zhang, Wang, Agarwal, Sajuyigbe, Chintala, Max, Chen, Kehoe, Satterfield, Govindaprasad, Gupta, Deng, Cho, Virk, Subramanian, Choudhury,
  Goldman, Remez, Glaser, Best, Koehler, Robinson, Li, Zhang, Matthews, Chou, Shaked, Vontimitta, Ajayi, Montanez, Mohan, Kumar, Mangla, Ionescu, Poenaru, Mihailescu, Ivanov, Li, Wang, Jiang, Bouaziz, Constable, Tang, Wu, Wang, Wu, Gao, Kleinman, Chen, Hu, Jia, Qi, Li, Zhang, Zhang, Adi, Nam, Yu, Wang, Zhao, Hao, Qian, Li, He, Rait, DeVito, Rosnbrick, Wen, Yang, Zhao, and Ma}]{grattafiori-etal-2024-llama3}
Aaron Grattafiori, Abhimanyu Dubey, Abhinav Jauhri, Abhinav Pandey, Abhishek Kadian, Ahmad Al-Dahle, Aiesha Letman, Akhil Mathur, Alan Schelten, Alex Vaughan, et~al. 2024.
\newblock \href {https://arxiv.org/abs/2407.21783} {The {Llama 3} herd of models}.
\newblock \emph{Preprint}, arXiv:2407.21783.

\bibitem[{Jiang et~al.(2020)Jiang, Xu, Araki, and Neubig}]{jiang-etal-2020-prompt-engineering}
Zhengbao Jiang, Frank~F. Xu, Jun Araki, and Graham Neubig. 2020.
\newblock \href {https://doi.org/10.1162/tacl_a_00324} {How can we know what language models know?}
\newblock \emph{Transactions of the Association for Computational Linguistics}, 8:423--438.

\bibitem[{Kaplan et~al.(2020)Kaplan, McCandlish, Henighan, Brown, Chess, Child, Gray, Radford, Wu, and Amodei}]{kaplan-etal-2020-scaling}
Jared Kaplan, Sam McCandlish, Tom Henighan, Tom~B. Brown, Benjamin Chess, Rewon Child, Scott Gray, Alec Radford, Jeffrey Wu, and Dario Amodei. 2020.
\newblock \href {https://arxiv.org/abs/2001.08361} {Scaling laws for neural language models}.
\newblock \emph{Preprint}, arXiv:2001.08361.

\bibitem[{Lai et~al.(2023)Lai, Ngo, Pouran Ben~Veyseh, Man, Dernoncourt, Bui, and Nguyen}]{lai-etal-2023-chatgpt}
Viet~Dac Lai, Nghia Ngo, Amir Pouran Ben~Veyseh, Hieu Man, Franck Dernoncourt, Trung Bui, and Thien~Huu Nguyen. 2023.
\newblock \href {https://doi.org/10.18653/v1/2023.findings-emnlp.878} {{C}hat{GPT} beyond {E}nglish: Towards a comprehensive evaluation of large language models in multilingual learning}.
\newblock In \emph{Findings of the Association for Computational Linguistics: EMNLP 2023}, pages 13171--13189, Singapore. Association for Computational Linguistics.

\bibitem[{Li et~al.(2023)Li, Koto, Wu, Aji, and Baldwin}]{li-etal-2023-bactrianx}
Haonan Li, Fajri Koto, Minghao Wu, Alham~Fikri Aji, and Timothy Baldwin. 2023.
\newblock \href {https://arxiv.org/abs/2305.15011} {{Bactrian-X}: Multilingual replicable instruction-following models with low-rank adaptation}.
\newblock \emph{Preprint}, arXiv:2305.15011.

\bibitem[{Lin et~al.(2024)Lin, Ji, Tiedemann, Martins, and Schütze}]{lin-etal-2024-mala500}
Peiqin Lin, Shaoxiong Ji, Jörg Tiedemann, André F.~T. Martins, and Hinrich Schütze. 2024.
\newblock \href {https://arxiv.org/abs/2401.13303} {{MaLA-500}: Massive language adaptation of large language models}.
\newblock \emph{Preprint}, arXiv:2401.13303.

\bibitem[{Lin et~al.(2022)Lin, Mihaylov, Artetxe, Wang, Chen, Simig, Ott, Goyal, Bhosale, Du, Pasunuru, Shleifer, Koura, Chaudhary, O{'}Horo, Wang, Zettlemoyer, Kozareva, Diab, Stoyanov, and Li}]{lin-etal-2022-shot}
Xi~Victoria Lin, Todor Mihaylov, Mikel Artetxe, Tianlu Wang, Shuohui Chen, Daniel Simig, Myle Ott, Naman Goyal, Shruti Bhosale, Jingfei Du, et~al. 2022.
\newblock \href {https://doi.org/10.18653/v1/2022.emnlp-main.616} {Few-shot learning with multilingual generative language models}.
\newblock In \emph{Proceedings of the 2022 Conference on Empirical Methods in Natural Language Processing}, pages 9019--9052, Abu Dhabi, United Arab Emirates. Association for Computational Linguistics.

\bibitem[{Liu et~al.(2022)Liu, Tam, Mohammed, Mohta, Huang, Bansal, and Raffel}]{liu-etal-2022-tfew}
Haokun Liu, Derek Tam, Muqeeth Mohammed, Jay Mohta, Tenghao Huang, Mohit Bansal, and Colin Raffel. 2022.
\newblock \href {https://openreview.net/forum?id=rBCvMG-JsPd} {Few-shot parameter-efficient fine-tuning is better and cheaper than in-context learning}.
\newblock In \emph{Advances in Neural Information Processing Systems}.

\bibitem[{Ma et~al.(2024)Ma, ImaniGooghari, Ye, Pei, Asgari, and Schütze}]{ma-etal-2024-taxi1500}
Chunlan Ma, Ayyoob ImaniGooghari, Haotian Ye, Renhao Pei, Ehsaneddin Asgari, and Hinrich Schütze. 2024.
\newblock \href {https://arxiv.org/abs/2305.08487} {{Taxi1500}: A multilingual dataset for text classification in 1500 languages}.
\newblock \emph{Preprint}, arXiv:2305.08487.

\bibitem[{Mart{\'i}nez-Garc{\'i}a et~al.(2021)Mart{\'i}nez-Garc{\'i}a, Badia, and Barnes}]{martinez-garcia-etal-2021-evaluating}
Antonio Mart{\'i}nez-Garc{\'i}a, Toni Badia, and Jeremy Barnes. 2021.
\newblock \href {https://doi.org/10.18653/v1/2021.acl-long.244} {Evaluating morphological typology in zero-shot cross-lingual transfer}.
\newblock In \emph{Proceedings of the 59th Annual Meeting of the Association for Computational Linguistics and the 11th International Joint Conference on Natural Language Processing (Volume 1: Long Papers)}, pages 3136--3153, Online. Association for Computational Linguistics.

\bibitem[{Martins et~al.(2025)Martins, Fernandes, Alves, Guerreiro, Rei, Alves, Pombal, Farajian, Faysse, Klimaszewski, Colombo, Haddow, {de Souza}, Birch, and Martins}]{matrins-etal-2025-eurollm}
Pedro~Henrique Martins, Patrick Fernandes, João Alves, Nuno~M. Guerreiro, Ricardo Rei, Duarte~M. Alves, José Pombal, Amin Farajian, Manuel Faysse, Mateusz Klimaszewski, et~al. 2025.
\newblock \href {https://doi.org/10.1016/j.procs.2025.02.260} {Eurollm: Multilingual language models for europe}.
\newblock \emph{Procedia Computer Science}, 255:53--62.
\newblock Proceedings of the Second EuroHPC user day.

\bibitem[{Micallef et~al.(2022)Micallef, Gatt, Tanti, van~der Plas, and Borg}]{micallef-etal-2022-pre}
Kurt Micallef, Albert Gatt, Marc Tanti, Lonneke van~der Plas, and Claudia Borg. 2022.
\newblock \href {https://doi.org/10.18653/v1/2022.deeplo-1.10} {Pre-training data quality and quantity for a low-resource language: New corpus and {BERT} models for {M}altese}.
\newblock In \emph{Proceedings of the Third Workshop on Deep Learning for Low-Resource Natural Language Processing}, pages 90--101, Hybrid. Association for Computational Linguistics.

\bibitem[{{Mistral {AI} Team}(2024)}]{mistral-2024-ministral}
{Mistral {AI} Team}. 2024.
\newblock Un {M}inistral, des {M}inistraux.
\newblock \url{https://mistral.ai/news/ministraux/}.
\newblock Accessed: 2024-12-20.

\bibitem[{Muennighoff et~al.(2023)Muennighoff, Wang, Sutawika, Roberts, Biderman, Le~Scao, Bari, Shen, Yong, Schoelkopf, Tang, Radev, Aji, Almubarak, Albanie, Alyafeai, Webson, Raff, and Raffel}]{muennighoff-etal-2023-crosslingual}
Niklas Muennighoff, Thomas Wang, Lintang Sutawika, Adam Roberts, Stella Biderman, Teven Le~Scao, M~Saiful Bari, Sheng Shen, Zheng~Xin Yong, Hailey Schoelkopf, et~al. 2023.
\newblock \href {https://doi.org/10.18653/v1/2023.acl-long.891} {Crosslingual generalization through multitask finetuning}.
\newblock In \emph{Proceedings of the 61st Annual Meeting of the Association for Computational Linguistics (Volume 1: Long Papers)}, pages 15991--16111, Toronto, Canada. Association for Computational Linguistics.

\bibitem[{Muller et~al.(2021)Muller, Anastasopoulos, Sagot, and Seddah}]{muller-etal-2021-unseen}
Benjamin Muller, Antonios Anastasopoulos, Beno{\^i}t Sagot, and Djam{\'e} Seddah. 2021.
\newblock \href {https://doi.org/10.18653/v1/2021.naacl-main.38} {When being unseen from m{BERT} is just the beginning: Handling new languages with multilingual language models}.
\newblock In \emph{Proceedings of the 2021 Conference of the North American Chapter of the Association for Computational Linguistics: Human Language Technologies}, pages 448--462, Online. Association for Computational Linguistics.

\bibitem[{{NLLB Team} et~al.(2022){NLLB Team}, Costa-jussà, Cross, Çelebi, Elbayad, Heafield, Heffernan, Kalbassi, Lam, Licht, Maillard, Sun, Wang, Wenzek, Youngblood, Akula, Barrault, Gonzalez, Hansanti, Hoffman, Jarrett, Sadagopan, Rowe, Spruit, Tran, Andrews, Ayan, Bhosale, Edunov, Fan, Gao, Goswami, Guzmán, Koehn, Mourachko, Ropers, Saleem, Schwenk, and Wang}]{nllb-2022-nllb}
{NLLB Team}, Marta~R. Costa-jussà, James Cross, Onur Çelebi, Maha Elbayad, Kenneth Heafield, Kevin Heffernan, Elahe Kalbassi, Janice Lam, Daniel Licht, et~al. 2022.
\newblock \href {https://arxiv.org/abs/2207.04672} {No language left behind: Scaling human-centered machine translation}.
\newblock \emph{Preprint}, arXiv:2207.04672.

\bibitem[{OpenAI et~al.(2024)OpenAI, Achiam, Adler, Agarwal, Ahmad, Akkaya, Aleman, Almeida, Altenschmidt, Altman, Anadkat, Avila, Babuschkin, Balaji, Balcom, Baltescu, Bao, Bavarian, Belgum, Bello, Berdine, Bernadett-Shapiro, Berner, Bogdonoff, Boiko, Boyd, Brakman, Brockman, Brooks, Brundage, Button, Cai, Campbell, Cann, Carey, Carlson, Carmichael, Chan, Chang, Chantzis, Chen, Chen, Chen, Chen, Chen, Chess, Cho, Chu, Chung, Cummings, Currier, Dai, Decareaux, Degry, Deutsch, Deville, Dhar, Dohan, Dowling, Dunning, Ecoffet, Eleti, Eloundou, Farhi, Fedus, Felix, Fishman, Forte, Fulford, Gao, Georges, Gibson, Goel, Gogineni, Goh, Gontijo-Lopes, Gordon, Grafstein, Gray, Greene, Gross, Gu, Guo, Hallacy, Han, Harris, He, Heaton, Heidecke, Hesse, Hickey, Hickey, Hoeschele, Houghton, Hsu, Hu, Hu, Huizinga, Jain, Jain, Jang, Jiang, Jiang, Jin, Jin, Jomoto, Jonn, Jun, Kaftan, Łukasz Kaiser, Kamali, Kanitscheider, Keskar, Khan, Kilpatrick, Kim, Kim, Kim, Kirchner, Kiros, Knight, Kokotajlo, Łukasz Kondraciuk,
  Kondrich, Konstantinidis, Kosic, Krueger, Kuo, Lampe, Lan, Lee, Leike, Leung, Levy, Li, Lim, Lin, Lin, Litwin, Lopez, Lowe, Lue, Makanju, Malfacini, Manning, Markov, Markovski, Martin, Mayer, Mayne, McGrew, McKinney, McLeavey, McMillan, McNeil, Medina, Mehta, Menick, Metz, Mishchenko, Mishkin, Monaco, Morikawa, Mossing, Mu, Murati, Murk, Mély, Nair, Nakano, Nayak, Neelakantan, Ngo, Noh, Ouyang, O'Keefe, Pachocki, Paino, Palermo, Pantuliano, Parascandolo, Parish, Parparita, Passos, Pavlov, Peng, Perelman, de~Avila Belbute~Peres, Petrov, de~Oliveira~Pinto, Michael, Pokorny, Pokrass, Pong, Powell, Power, Power, Proehl, Puri, Radford, Rae, Ramesh, Raymond, Real, Rimbach, Ross, Rotsted, Roussez, Ryder, Saltarelli, Sanders, Santurkar, Sastry, Schmidt, Schnurr, Schulman, Selsam, Sheppard, Sherbakov, Shieh, Shoker, Shyam, Sidor, Sigler, Simens, Sitkin, Slama, Sohl, Sokolowsky, Song, Staudacher, Such, Summers, Sutskever, Tang, Tezak, Thompson, Tillet, Tootoonchian, Tseng, Tuggle, Turley, Tworek, Uribe, Vallone,
  Vijayvergiya, Voss, Wainwright, Wang, Wang, Wang, Ward, Wei, Weinmann, Welihinda, Welinder, Weng, Weng, Wiethoff, Willner, Winter, Wolrich, Wong, Workman, Wu, Wu, Wu, Xiao, Xu, Yoo, Yu, Yuan, Zaremba, Zellers, Zhang, Zhang, Zhao, Zheng, Zhuang, Zhuk, and Zoph}]{openai-2024-gpt4}
OpenAI, Josh Achiam, Steven Adler, Sandhini Agarwal, Lama Ahmad, Ilge Akkaya, Florencia~Leoni Aleman, Diogo Almeida, Janko Altenschmidt, Sam Altman, et~al. 2024.
\newblock \href {https://arxiv.org/abs/2303.08774} {{GPT-4} technical report}.
\newblock \emph{Preprint}, arXiv:2303.08774.

\bibitem[{Rafailov et~al.(2023)Rafailov, Sharma, Mitchell, Manning, Ermon, and Finn}]{rafailov-etal-2023-dpo}
Rafael Rafailov, Archit Sharma, Eric Mitchell, Christopher~D Manning, Stefano Ermon, and Chelsea Finn. 2023.
\newblock \href {https://openreview.net/forum?id=HPuSIXJaa9} {Direct preference optimization: Your language model is secretly a reward model}.
\newblock In \emph{Thirty-seventh Conference on Neural Information Processing Systems}.

\bibitem[{Raffel et~al.(2020)Raffel, Shazeer, Roberts, Lee, Narang, Matena, Zhou, Li, and Liu}]{raffel-etal-2020-t5}
Colin Raffel, Noam Shazeer, Adam Roberts, Katherine Lee, Sharan Narang, Michael Matena, Yanqi Zhou, Wei Li, and Peter~J. Liu. 2020.
\newblock \href {http://jmlr.org/papers/v21/20-074.html} {Exploring the limits of transfer learning with a unified text-to-text transformer}.
\newblock \emph{Journal of Machine Learning Research}, 21(140):1--67.

\bibitem[{Rosner and Borg(2023)}]{rosner-borg-2022-ele_maltese}
Michael Rosner and Claudia Borg. 2023.
\newblock \href {https://doi.org/10.1007/978-3-031-28819-7_27} {{Language Report Maltese}}.
\newblock In Georg Rehm and Andy Way, editors, \emph{European Language Equality: A Strategic Agenda for Digital Language Equality}, pages 183--186. Springer International Publishing, Cham.

\bibitem[{Shin et~al.(2020)Shin, Razeghi, Logan~IV, Wallace, and Singh}]{shin-etal-2020-autoprompt}
Taylor Shin, Yasaman Razeghi, Robert~L. Logan~IV, Eric Wallace, and Sameer Singh. 2020.
\newblock \href {https://doi.org/10.18653/v1/2020.emnlp-main.346} {{A}uto{P}rompt: {E}liciting {K}nowledge from {L}anguage {M}odels with {A}utomatically {G}enerated {P}rompts}.
\newblock In \emph{Proceedings of the 2020 Conference on Empirical Methods in Natural Language Processing (EMNLP)}, pages 4222--4235, Online. Association for Computational Linguistics.

\bibitem[{Shliazhko et~al.(2024)Shliazhko, Fenogenova, Tikhonova, Kozlova, Mikhailov, and Shavrina}]{shliazhko-etal-2024-mgpt}
Oleh Shliazhko, Alena Fenogenova, Maria Tikhonova, Anastasia Kozlova, Vladislav Mikhailov, and Tatiana Shavrina. 2024.
\newblock \href {https://doi.org/10.1162/tacl_a_00633} {m{GPT}: Few-shot learners go multilingual}.
\newblock \emph{Transactions of the Association for Computational Linguistics}, 12:58--79.

\bibitem[{Touvron et~al.(2023)Touvron, Martin, Stone, Albert, Almahairi, Babaei, Bashlykov, Batra, Bhargava, Bhosale, Bikel, Blecher, Ferrer, Chen, Cucurull, Esiobu, Fernandes, Fu, Fu, Fuller, Gao, Goswami, Goyal, Hartshorn, Hosseini, Hou, Inan, Kardas, Kerkez, Khabsa, Kloumann, Korenev, Koura, Lachaux, Lavril, Lee, Liskovich, Lu, Mao, Martinet, Mihaylov, Mishra, Molybog, Nie, Poulton, Reizenstein, Rungta, Saladi, Schelten, Silva, Smith, Subramanian, Tan, Tang, Taylor, Williams, Kuan, Xu, Yan, Zarov, Zhang, Fan, Kambadur, Narang, Rodriguez, Stojnic, Edunov, and Scialom}]{touvron-etal-2023-llama2}
Hugo Touvron, Louis Martin, Kevin Stone, Peter Albert, Amjad Almahairi, Yasmine Babaei, Nikolay Bashlykov, Soumya Batra, Prajjwal Bhargava, Shruti Bhosale, et~al. 2023.
\newblock \href {https://arxiv.org/abs/2307.09288} {{Llama 2}: Open foundation and fine-tuned chat models}.
\newblock \emph{Preprint}, arXiv:2307.09288.

\bibitem[{{\"U}st{\"u}n et~al.(2024){\"U}st{\"u}n, Aryabumi, Yong, Ko, D{'}souza, Onilude, Bhandari, Singh, Ooi, Kayid, Vargus, Blunsom, Longpre, Muennighoff, Fadaee, Kreutzer, and Hooker}]{ustun-etal-2024-aya}
Ahmet {\"U}st{\"u}n, Viraat Aryabumi, Zheng Yong, Wei-Yin Ko, Daniel D{'}souza, Gbemileke Onilude, Neel Bhandari, Shivalika Singh, Hui-Lee Ooi, Amr Kayid, et~al. 2024.
\newblock \href {https://doi.org/10.18653/v1/2024.acl-long.845} {Aya model: An instruction finetuned open-access multilingual language model}.
\newblock In \emph{Proceedings of the 62nd Annual Meeting of the Association for Computational Linguistics (Volume 1: Long Papers)}, pages 15894--15939, Bangkok, Thailand. Association for Computational Linguistics.

\bibitem[{Wei et~al.(2023)Wei, Wei, Lin, Li, Zhang, Ren, Li, Wan, Cao, Xie, Hu, Li, Hui, Yu, Liu, Yang, Huang, and Xie}]{wei-etal-2023-polylm}
Xiangpeng Wei, Haoran Wei, Huan Lin, Tianhao Li, Pei Zhang, Xingzhang Ren, Mei Li, Yu~Wan, Zhiwei Cao, Binbin Xie, et~al. 2023.
\newblock \href {https://arxiv.org/abs/2307.06018} {{PolyLM}: An open source polyglot large language model}.
\newblock \emph{Preprint}, arXiv:2307.06018.

\bibitem[{Wolf et~al.(2020)Wolf, Debut, Sanh, Chaumond, Delangue, Moi, Cistac, Rault, Louf, Funtowicz, Davison, Shleifer, von Platen, Ma, Jernite, Plu, Xu, Le~Scao, Gugger, Drame, Lhoest, and Rush}]{wolf-etal-2020-transformers}
Thomas Wolf, Lysandre Debut, Victor Sanh, Julien Chaumond, Clement Delangue, Anthony Moi, Pierric Cistac, Tim Rault, Remi Louf, Morgan Funtowicz, et~al. 2020.
\newblock \href {https://doi.org/10.18653/v1/2020.emnlp-demos.6} {Transformers: State-of-the-art natural language processing}.
\newblock In \emph{Proceedings of the 2020 Conference on Empirical Methods in Natural Language Processing: System Demonstrations}, pages 38--45, Online. Association for Computational Linguistics.

\bibitem[{Xue et~al.(2021)Xue, Constant, Roberts, Kale, Al-Rfou, Siddhant, Barua, and Raffel}]{xue-etal-2021-mt5}
Linting Xue, Noah Constant, Adam Roberts, Mihir Kale, Rami Al-Rfou, Aditya Siddhant, Aditya Barua, and Colin Raffel. 2021.
\newblock \href {https://doi.org/10.18653/v1/2021.naacl-main.41} {m{T}5: A massively multilingual pre-trained text-to-text transformer}.
\newblock In \emph{Proceedings of the 2021 Conference of the North American Chapter of the Association for Computational Linguistics: Human Language Technologies}, pages 483--498, Online. Association for Computational Linguistics.

\bibitem[{Zhang et~al.(2024)Zhang, Gautam, Wang, Alabi, Shen, Klakow, and Mosbach}]{zhang-etal-2024-impact}
Miaoran Zhang, Vagrant Gautam, Mingyang Wang, Jesujoba Alabi, Xiaoyu Shen, Dietrich Klakow, and Marius Mosbach. 2024.
\newblock \href {https://doi.org/10.18653/v1/2024.findings-acl.438} {The impact of demonstrations on multilingual in-context learning: A multidimensional analysis}.
\newblock In \emph{Findings of the Association for Computational Linguistics: ACL 2024}, pages 7342--7371, Bangkok, Thailand. Association for Computational Linguistics.

\bibitem[{Zhang et~al.(2023)Zhang, Cahyawijaya, Cruz, Winata, and Aji}]{zhang-etal-2023-multilingual}
Ruochen Zhang, Samuel Cahyawijaya, Jan Christian~Blaise Cruz, Genta Winata, and Alham~Fikri Aji. 2023.
\newblock \href {https://doi.org/10.18653/v1/2023.emnlp-main.774} {Multilingual large language models are not (yet) code-switchers}.
\newblock In \emph{Proceedings of the 2023 Conference on Empirical Methods in Natural Language Processing}, pages 12567--12582, Singapore. Association for Computational Linguistics.

\bibitem[{Zhao et~al.(2021)Zhao, Wallace, Feng, Klein, and Singh}]{zhao-etal-2021-calibrate}
Zihao Zhao, Eric Wallace, Shi Feng, Dan Klein, and Sameer Singh. 2021.
\newblock \href {https://proceedings.mlr.press/v139/zhao21c.html} {Calibrate before use: Improving few-shot performance of language models}.
\newblock In \emph{Proceedings of the 38th International Conference on Machine Learning}, volume 139 of \emph{Proceedings of Machine Learning Research}, pages 12697--12706. PMLR.

\bibitem[{Ziegler et~al.(2020)Ziegler, Stiennon, Wu, Brown, Radford, Amodei, Christiano, and Irving}]{ziegler-etal-2020-rlhf}
Daniel~M. Ziegler, Nisan Stiennon, Jeffrey Wu, Tom~B. Brown, Alec Radford, Dario Amodei, Paul Christiano, and Geoffrey Irving. 2020.
\newblock \href {https://arxiv.org/abs/1909.08593} {Fine-tuning language models from human preferences}.
\newblock \emph{Preprint}, arXiv:1909.08593.

\end{thebibliography}

\newpage
\appendix

\section{Prompt Templates}
\label{appendix:prompts}

Table~\ref{table:prompts} shows the prompts that were used to evaluate every model on each task.
When available, we use a template suggested by the original dataset paper.
Otherwise, we adapt a template from a related task which was available from the Language Model Evaluation Harness repository \cite{lm-evaluation-harness}.
For \NLG{} tasks, we ensure that the instruction explicitly mentions that the text should be generated in Maltese.

To generate the Maltese prompts, we reuse the English templates and translate them into Maltese.
This is done by first passing the instruction through Google Translate and then performing post-editing with a Maltese native speaker.

\begin{table*}[t!]
    \centering
    \begin{tabular}{|p{2.5cm}||p{6cm}|p{6cm}|}
        \hline
        Task & English Prompt Template & Maltese Prompt Template \\
        \hline\hline
        Sentiment Analysis & \texttt{\{text\}} Is the sentiment positive or negative? & \texttt{\{text\}} Is-sentiment huwa pożittiv jew negattiv? \\
        \hline
        SIB-200 & The topic of the news ``\texttt{\{text\}}'' is & Is-suġġett tal-a\mh barjiet ``\texttt{\{text\}}'' huwa \\
        \hline
        Taxi1500 & The topic of the verse is ``\texttt{\{text\}}'' is & Is-suġġett tal-vers ``\texttt{\{text\}}'' huwa \\
        \hline
        Maltese News Categories & \texttt{\{text\}}\newline\newline What are the topic(s) of this news article? & \texttt{\{text\}}\newline\newline X'inhu(ma) s-suġġett(i) ta' dan l-artiklu tal-a\mh barjiet? \\
        \hline
        MultiEURLEX & \texttt{\{text\}}\newline\newline What are the topics of this text? & \texttt{\{text\}}\newline\newline X'inhuma s-suġġetti ta' dan it-test? \\
        \hline
        Belebele & Given the following passage, query, and answer choices, output the letter corresponding to the correct answer.\newline \#\#\#\newline Passage:\newline \texttt{\{text\}}\newline \#\#\#\newline Query:\newline \texttt{\{question\}}\newline \#\#\#\newline Choices:\newline (A) \texttt{\{answer1\}}\newline (B) \texttt{\{answer2\}}\newline (C) \texttt{\{answer3\}}\newline (D) \texttt{\{answer4\}}\newline \#\#\#\newline Answer:\newline & Permezz tas-silta, mistoqsija, u g\mh ażliet ta' tweġibiet li ġejjin, ag\mh ti l-ittra li tikkorrispondi g\mh at-tweġiba t-tajba.\newline \#\#\#\newline Passaġġ:\newline \texttt{\{text\}}\newline \#\#\#\newline Mistoqsija:\newline \texttt{\{question\}}\newline \#\#\#\newline Choices:\newline (A) \texttt{\{answer1\}}\newline (B) \texttt{\{answer2\}}\newline (C) \texttt{\{answer3\}}\newline (D) \texttt{\{answer4\}}\newline \#\#\#\newline Tweġiba:\newline \\
        \hline
        OPUS-100 \newline Flores-200 & \texttt{\{source\_sentence\}}\newline\newline The previous text is in \texttt{\{source\_language\}}. Here is a translation to \texttt{\{target\_language\}}: & \texttt{\{source\_sentence\}}\newline\newline It-test preċedenti huwa bl-\texttt{\{source\_language\}}. Din hija traduzzjoni g\mh all-\texttt{\{target\_language\}}: \\
        \hline
        WebNLG & Verbalize in Maltese the following triples separated by a comma: \texttt{\{triples | join(', ')\}} & Ivverbalizza bil-Malti t-tripli li ġejjin separati b'virgola: \texttt{\{triples | join(', ')\}} \\
        \hline
        EUR-Lex-Sum & \texttt{\{text\}}\newline\newline Write a summary in Maltese for the text above: & \texttt{\{text\}}\newline\newline Ikteb sommarju bil-Malti g\mh at-test t'hawn fuq:\\
        \hline
        Maltese News Headlines & \texttt{\{text\}}\newline\newline Write a headline in Maltese for the news article above: & \texttt{\{text\}}\newline\newline Ikteb titolu bil-Malti g\mh all-artiklu tal-a\mh barjiet t'hawn fuq: \\
        \hline
    \end{tabular}
    \caption{Prompt Templates used for each task.}
    \label{table:prompts}
\end{table*}

\section{Fine-Tuning Details}

\label{appendix:experiments_finetuning}

For fine-tuned models, we consider BERTu \cite{micallef-etal-2022-pre}, mBERT \cite{devlin-etal-2019-bert}, and mT5-Small \cite{xue-etal-2021-mt5}.
Our training scripts are implemented using the \texttt{transformers} library \cite{wolf-etal-2020-transformers}.

We train all models for at most 200 epochs but use early stopping on the main metric (as defined in Section~\ref{section:evaluation}) with a patience of 20 epochs.
Due to the significantly larger scale of the data for OPUS-100, we only train for a maximum of 10 epochs instead.

For the BERT-based models, we use a learning rate of 1e-4 with an AdamW optimiser, an inverse square-root learning rate scheduler, a warmup of 1 epoch, and a weight decay of 0.01.
We use a batch size of 16 for Sentiment Analysis, SIB-200, and Taxi1500 and a batch size of 32 for the other \NLU{} tasks.
A dropout of 0.1 is used for Taxi1500 and MultiEURLEX, and 0.5 for the other \NLU{} tasks.

When fine-tuning mT5, we mostly follow the original fine-tuning recipe \cite{raffel-etal-2020-t5, xue-etal-2021-mt5} with a constant learning rate of 1e-3 with an Adafactor optimiser and a batch size of 32.
BERT-based models are fine-tuned 5 separate times with different random seeds, and we report the mean performance across these runs.
mT5 is only fine-tuned once per task.

\section{All Results}
\label{appendix:all_results}

\begin{figure*}[t!]
    \centering
    \includegraphics[scale=0.6]{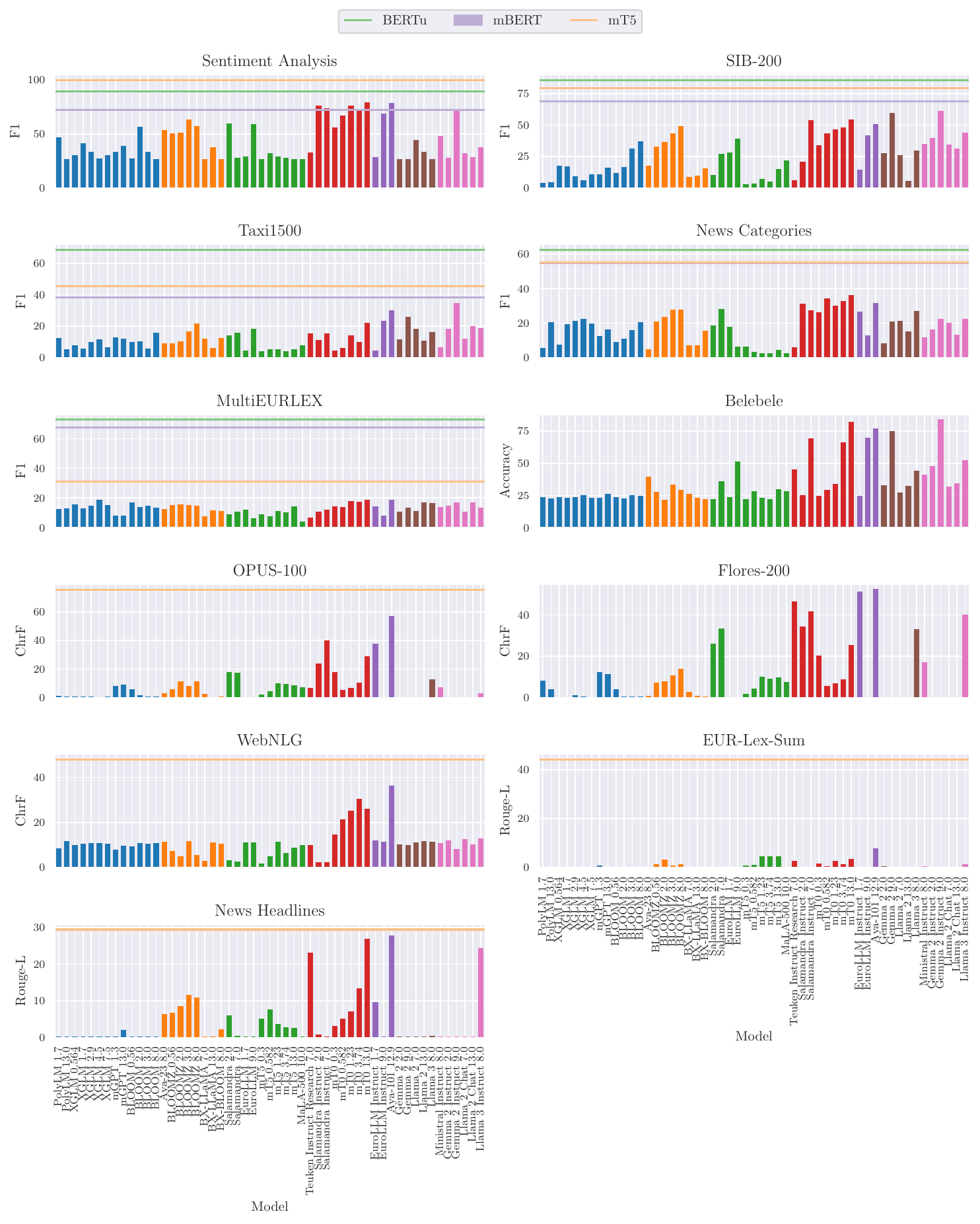}
    \caption{Zero-shot prompting performance of individual models on each task. Horizontal lines represent models fine-tuned specifically on the task.}
    \label{figure:model_task_0shot_performance}
\end{figure*}

\begin{figure*}[t!]
    \centering
    \includegraphics[scale=0.6]{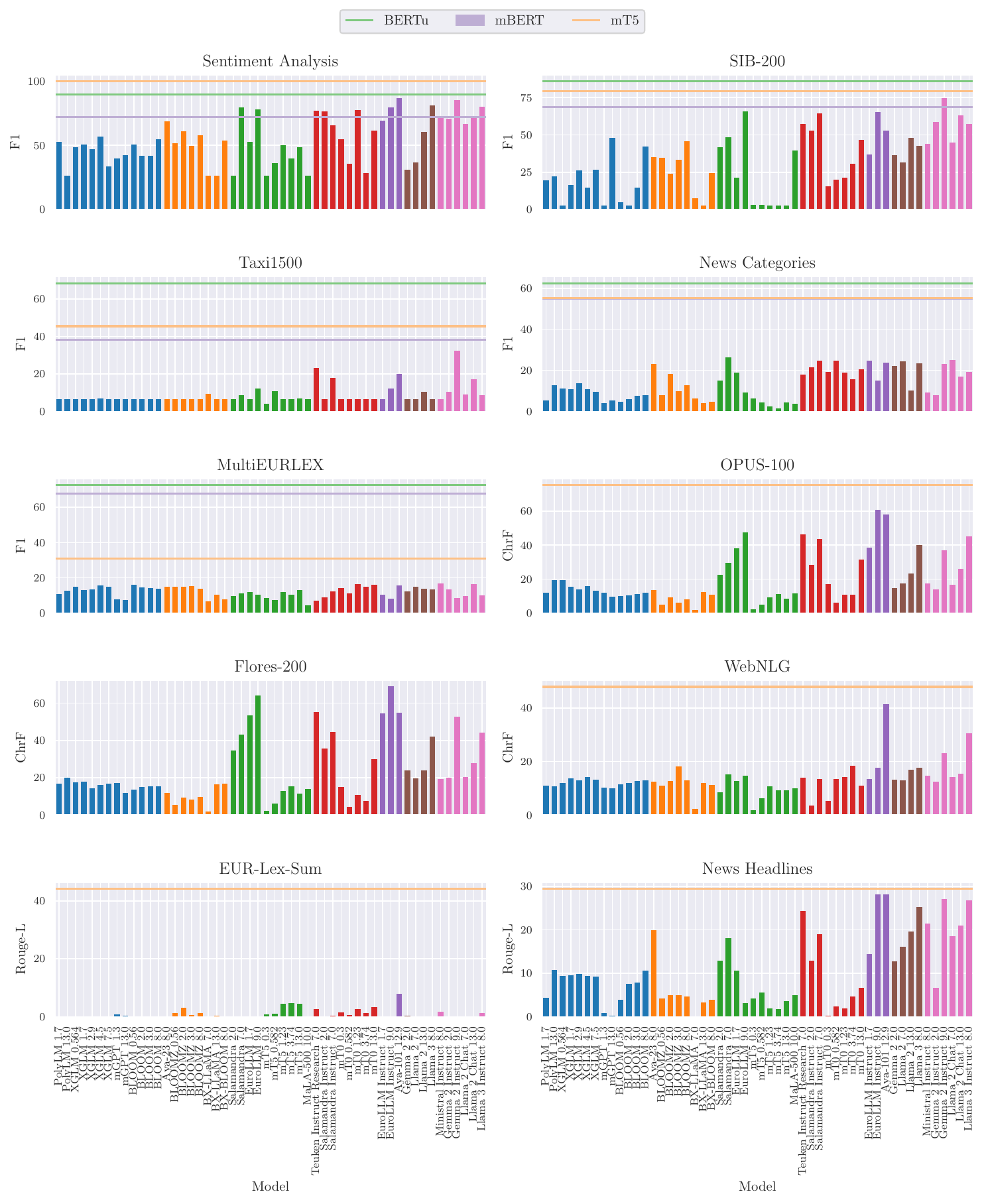}
    \caption{One-shot prompting performance of individual models on each task. Horizontal lines represent models fine-tuned specifically on the task.}
    \label{figure:model_task_1shot_performance}
\end{figure*}

In this section, we present individual results for each model and task considered.
Figure~\ref{figure:model_task_0shot_performance} shows the zero-shot performance with English instructions and Figure~\ref{figure:model_task_1shot_performance} shows the one-shot performance with English instructions.

We also present the individual performance with all metrics of each model with English prompts in Tables~\ref{table:results_nlu_0shot}, \ref{table:results_nlg_0shot}, \ref{table:results_nlu_1shot}, and~\ref{table:results_nlg_1shot}.
Results for fine-tuned models are shown in Tables~\ref{table:results_nlu_finetuned} and~\ref{table:results_nlg_finetuned}.

\begin{table*}[t]
    \footnotesize
    \setlength{\tabcolsep}{3pt}
    \centering
    \begin{tabular}{|l||c|c|c|c|c|c|}
        \hline
        \multirow{2}{*}{\textbf{Model}} & \textbf{Sentiment Analysis} & \textbf{SIB-200} & \textbf{Taxi1500} & \textbf{News Categories} & \textbf{MultiEURLEX} & \textbf{Belebele} \\
         & F1 & F1 & F1 & F1 & F1 & Accuracy \\
        \hline\hline
        PolyLM\textsubscript{1.7B}	 & 46.3 & ~3.4 & 12.0 & ~5.4 & 12.1 & 23.0 \\
        PolyLM\textsubscript{13.0B}	 & 26.0 & ~3.8 & ~5.0 & 20.4 & 12.6 & 22.2 \\
        XGLM\textsubscript{0.564B}	 & 29.4 & 17.1 & ~7.5 & ~7.1 & 15.1 & 23.1 \\
        XGLM\textsubscript{1.7B}	 & 40.9 & 16.6 & ~5.5 & 18.9 & 12.7 & 23.0 \\
        XGLM\textsubscript{2.9B}	 & 32.7 & ~8.9 & ~9.7 & 21.0 & 14.3 & 23.4 \\
        XGLM\textsubscript{4.5B}	 & 26.6 & ~5.8 & 11.4 & 22.1 & 18.2 & 24.7 \\
        XGLM\textsubscript{7.5B}	 & 29.9 & 10.2 & ~6.4 & 19.3 & 14.9 & 22.9 \\
        mGPT\textsubscript{1.3B}	 & 33.0 & 10.1 & 12.5 & 12.0 & ~7.9 & 22.7 \\
        mGPT\textsubscript{13.0B}	 & 38.5 & 15.7 & 11.7 & 16.0 & ~7.5 & 25.7 \\
        BLOOM\textsubscript{0.56B}	 & 26.6 & 11.1 & ~9.6 & ~8.9 & 16.4 & 23.1 \\
        BLOOM\textsubscript{2.0B}	 & 56.1 & 15.9 & 10.0 & 10.7 & 13.7 & 22.2 \\
        BLOOM\textsubscript{3.0B}	 & 32.9 & 30.6 & ~5.3 & 15.6 & 14.4 & 24.8 \\
        BLOOM\textsubscript{8.0B}	 & 26.0 & 36.6 & 15.3 & 20.0 & 12.8 & 24.1 \\
        Aya-23\textsubscript{8.0B}	 & 53.1 & 16.9 & ~8.9 & ~4.5 & 12.2 & 39.3 \\
        BLOOMZ\textsubscript{0.56B}	 & 49.9 & 32.6 & ~8.9 & 20.8 & 14.9 & 27.4 \\
        BLOOMZ\textsubscript{2.0B}	 & 50.4 & 36.0 & ~9.8 & 23.3 & 15.1 & 21.0 \\
        BLOOMZ\textsubscript{3.0B}	 & 62.8 & 42.9 & 16.3 & 27.5 & 14.9 & 32.9 \\
        BLOOMZ\textsubscript{8.0B}	 & 56.4 & 48.8 & 21.2 & 27.4 & 14.2 & 28.9 \\
        BX-LLaMA\textsubscript{7.0B}	 & 26.0 & ~8.0 & 11.6 & ~6.9 & ~7.4 & 25.7 \\
        BX-LLaMA\textsubscript{13.0B}	 & 36.7 & ~9.1 & ~5.7 & ~7.0 & 11.3 & 22.8 \\
        BX-BLOOM\textsubscript{8.0B}	 & 26.0 & 14.8 & 12.1 & 15.2 & 10.8 & 21.7 \\
        \hline
        Salamandra\textsubscript{2.0B}	 & 59.0 & ~9.8 & 13.6 & 18.3 & ~8.6 & 21.6 \\
        Salamandra\textsubscript{7.0B}	 & 27.0 & 26.4 & 15.4 & 27.7 & 10.5 & 35.7 \\
        EuroLLM\textsubscript{1.7B}	 & 28.6 & 27.5 & ~4.2 & 17.4 & 11.9 & 23.1 \\
        EuroLLM\textsubscript{9.0B}	 & 58.5 & 38.7 & 18.1 & ~6.1 & ~5.9 & 51.2 \\
        mT5\textsubscript{0.3B}	 & 26.0 & ~2.4 & ~3.7 & ~6.2 & ~8.7 & 21.9 \\
        mT5\textsubscript{0.582B}	 & 31.6 & ~2.8 & ~4.9 & ~2.9 & ~7.3 & 27.9 \\
        mT5\textsubscript{1.23B}	 & 28.2 & ~6.7 & ~4.9 & ~2.1 & 11.0 & 22.9 \\
        mT5\textsubscript{3.74B}	 & 27.0 & ~4.3 & ~3.7 & ~2.1 & ~9.9 & 21.9 \\
        mT5\textsubscript{13.0B}	 & 26.0 & 14.6 & ~4.9 & ~4.1 & 14.0 & 29.3 \\
        MaLA-500\textsubscript{10.0B}	 & 26.0 & 21.4 & ~7.4 & ~2.3 & ~3.6 & 27.9 \\
        Teuken Instruct Research\textsubscript{7.0B}	 & 32.1 & ~5.7 & 15.2 & ~5.7 & ~6.2 & 45.0 \\
        Salamandra Instruct\textsubscript{2.0B}	 & 75.4 & 20.0 & 10.7 & 31.0 & 10.3 & 24.9 \\
        Salamandra Instruct\textsubscript{7.0B}	 & 73.4 & 53.2 & 15.1 & 26.9 & 11.7 & 69.0 \\
        mT0\textsubscript{0.3B}	 & 55.5 & 33.6 & ~4.3 & 25.8 & 13.8 & 24.4 \\
        mT0\textsubscript{0.582B}	 & 66.3 & 43.1 & ~5.9 & 33.8 & 13.4 & 28.9 \\
        mT0\textsubscript{1.23B}	 & 75.9 & 45.8 & 13.6 & 29.8 & 17.6 & 33.7 \\
        mT0\textsubscript{3.74B}	 & 72.3 & 47.5 & ~9.6 & 32.5 & 16.8 & 65.9 \\
        mT0\textsubscript{13.0B}	 & 78.5 & 54.2 & 21.7 & 36.1 & 18.3 & 81.7 \\
        \hline
        EuroLLM Instruct\textsubscript{1.7B}	 & 28.0 & 14.0 & ~4.2 & 26.3 & 13.7 & 24.1 \\
        EuroLLM Instruct\textsubscript{9.0B}	 & 68.1 & 41.2 & 23.0 & 12.6 & ~7.7 & 69.6 \\
        Aya-101\textsubscript{12.9B}	 & 78.1 & 50.2 & 29.5 & 31.5 & 18.5 & 76.6 \\
        \hline
        Gemma 2\textsubscript{2.0B}	 & 26.0 & 27.0 & 11.2 & ~7.9 & 10.2 & 32.7 \\
        Gemma 2\textsubscript{9.0B}	 & 26.0 & 59.2 & 25.5 & 20.7 & 13.2 & 74.6 \\
        Llama 2\textsubscript{7.0B}	 & 43.8 & 25.8 & 17.9 & 21.0 & 10.7 & 26.9 \\
        Llama 2\textsubscript{13.0B}	 & 32.9 & ~5.1 & 10.6 & 15.0 & 16.4 & 31.8 \\
        Llama 3\textsubscript{8.0B}	 & 26.0 & 29.5 & 15.7 & 26.8 & 16.2 & 43.7 \\
        Ministral Instruct\textsubscript{8.0B}	 & 47.4 & 34.4 & ~6.2 & 11.3 & 13.4 & 40.6 \\
        Gemma 2 Instruct\textsubscript{2.0B}	 & 27.0 & 39.4 & 18.2 & 15.8 & 14.4 & 47.6 \\
        Gemma 2 Instruct\textsubscript{9.0B}	 & 72.1 & 60.9 & 34.2 & 22.3 & 16.4 & 83.9 \\
        Llama 2 Chat\textsubscript{7.0B}	 & 31.8 & 34.0 & 11.9 & 19.9 & 10.3 & 31.6 \\
        Llama 2 Chat\textsubscript{13.0B}	 & 28.0 & 30.9 & 19.6 & 13.1 & 16.6 & 34.0 \\
        Llama 3 Instruct\textsubscript{8.0B}	 & 37.3 & 43.6 & 18.4 & 22.0 & 13.2 & 51.9 \\
        \hline
    \end{tabular}
    \caption{Results on \NLU{} tasks for models prompted with English zero-shot instructions.}
    \label{table:results_nlu_0shot}
\end{table*}

\begin{table*}[t]
    \footnotesize
    \setlength{\tabcolsep}{3pt}
    \centering
    \begin{tabular}{|l||cc|cc|cc|cc|cc|}
        \hline
        \multirow{2}{*}{\textbf{Model}} & \multicolumn{2}{c|}{\textbf{OPUS-100}} & \multicolumn{2}{c|}{\textbf{Flores-200}} & \multicolumn{2}{c|}{\textbf{WebNLG}} & \multicolumn{2}{c|}{\textbf{EUR-Lex-Sum}} & \multicolumn{2}{c|}{\textbf{News Headlines}} \\
         & BLEU & ChrF & BLEU & ChrF & ChrF & Rouge-L & ChrF & Rouge-L & ChrF & Rouge-L \\
        \hline\hline
        PolyLM\textsubscript{1.7B}	 & ~0.0 & ~0.6 & ~0.1 & ~7.7 & ~8.2 & ~3.7 & ~0.0 & ~0.0 & ~0.0 & ~0.0 \\
        PolyLM\textsubscript{13.0B}	 & ~0.0 & ~0.4 & ~0.1 & ~3.7 & 11.5 & ~6.0 & ~0.0 & ~0.0 & ~0.0 & ~0.0 \\
        XGLM\textsubscript{0.564B}	 & ~0.0 & ~0.1 & ~0.0 & ~0.0 & ~9.5 & ~4.3 & ~0.0 & ~0.0 & ~0.0 & ~0.0 \\
        XGLM\textsubscript{1.7B}	 & ~0.0 & ~0.1 & ~0.0 & ~0.0 & 10.1 & ~4.9 & ~0.0 & ~0.0 & ~0.0 & ~0.0 \\
        XGLM\textsubscript{2.9B}	 & ~0.0 & ~0.1 & ~0.0 & ~0.9 & 10.6 & ~4.9 & ~0.0 & ~0.0 & ~0.4 & ~0.1 \\
        XGLM\textsubscript{4.5B}	 & ~0.0 & ~0.0 & ~0.0 & ~0.1 & 10.6 & ~4.8 & ~0.0 & ~0.0 & ~0.1 & ~0.0 \\
        XGLM\textsubscript{7.5B}	 & ~0.0 & ~0.1 & ~0.0 & ~0.0 & 10.2 & ~4.8 & ~0.0 & ~0.0 & ~0.2 & ~0.1 \\
        mGPT\textsubscript{1.3B}	 & ~0.2 & ~7.8 & ~0.3 & 11.9 & ~7.7 & ~3.0 & ~0.2 & ~0.6 & ~0.4 & ~0.1 \\
        mGPT\textsubscript{13.0B}	 & ~0.3 & ~8.5 & ~0.2 & 11.0 & ~9.3 & ~4.3 & ~0.1 & ~0.1 & ~5.0 & ~1.9 \\
        BLOOM\textsubscript{0.56B}	 & ~0.2 & ~5.5 & ~0.3 & ~3.9 & ~9.1 & ~4.6 & ~0.0 & ~0.0 & ~0.0 & ~0.0 \\
        BLOOM\textsubscript{2.0B}	 & ~0.1 & ~1.1 & ~0.0 & ~0.1 & 10.5 & ~5.4 & ~0.0 & ~0.0 & ~0.0 & ~0.0 \\
        BLOOM\textsubscript{3.0B}	 & ~0.0 & ~0.4 & ~0.0 & ~0.3 & 10.4 & ~5.4 & ~0.0 & ~0.0 & ~0.0 & ~0.0 \\
        BLOOM\textsubscript{8.0B}	 & ~0.0 & ~0.2 & ~0.0 & ~0.2 & 10.6 & ~5.3 & ~0.0 & ~0.0 & ~0.0 & ~0.0 \\
        Aya-23\textsubscript{8.0B}	 & ~0.1 & ~2.6 & ~0.0 & ~0.5 & 11.2 & ~4.9 & ~0.0 & ~0.0 & 11.5 & ~6.2 \\
        BLOOMZ\textsubscript{0.56B}	 & ~0.5 & ~5.3 & ~0.2 & ~6.8 & ~7.1 & ~5.0 & ~0.7 & ~1.0 & 12.9 & ~6.6 \\
        BLOOMZ\textsubscript{2.0B}	 & ~1.2 & 10.9 & ~0.2 & ~7.7 & ~4.7 & ~3.4 & ~2.0 & ~2.9 & 14.3 & ~8.3 \\
        BLOOMZ\textsubscript{3.0B}	 & ~0.7 & ~7.8 & ~0.5 & 10.3 & 11.5 & 10.2 & ~0.4 & ~0.5 & 17.5 & 11.4 \\
        BLOOMZ\textsubscript{8.0B}	 & ~1.4 & 11.0 & ~1.0 & 13.7 & ~5.3 & ~4.6 & ~0.8 & ~1.1 & 16.8 & 10.7 \\
        BX-LLaMA\textsubscript{7.0B}	 & ~0.0 & ~2.0 & ~0.0 & ~2.4 & ~2.6 & ~0.8 & ~1.2 & ~0.0 & ~1.3 & ~0.1 \\
        BX-LLaMA\textsubscript{13.0B}	 & ~0.0 & ~0.0 & ~0.0 & ~0.6 & 10.9 & ~5.5 & ~0.1 & ~0.1 & ~0.8 & ~0.1 \\
        BX-BLOOM\textsubscript{8.0B}	 & ~0.0 & ~0.3 & ~0.0 & ~0.2 & 10.3 & ~5.1 & ~0.0 & ~0.0 & ~5.9 & ~2.1 \\
        \hline
        Salamandra\textsubscript{2.0B}	 & ~1.4 & 17.3 & ~3.7 & 25.7 & ~2.9 & ~2.5 & ~0.0 & ~0.0 & 10.9 & ~5.8 \\
        Salamandra\textsubscript{7.0B}	 & ~2.0 & 17.1 & ~7.9 & 33.0 & ~2.4 & ~1.6 & ~0.0 & ~0.0 & ~0.3 & ~0.2 \\
        EuroLLM\textsubscript{1.7B}	 & ~0.0 & ~0.0 & ~0.0 & ~0.0 & 10.9 & ~5.5 & ~0.0 & ~0.0 & ~0.0 & ~0.0 \\
        EuroLLM\textsubscript{9.0B}	 & ~0.0 & ~0.0 & ~0.0 & ~0.0 & 10.8 & ~4.5 & ~0.0 & ~0.0 & ~0.0 & ~0.0 \\
        mT5\textsubscript{0.3B}	 & ~0.1 & ~1.5 & ~0.0 & ~1.5 & ~1.4 & ~0.6 & ~0.2 & ~0.5 & ~6.3 & ~4.9 \\
        mT5\textsubscript{0.582B}	 & ~0.2 & ~3.9 & ~0.0 & ~4.1 & ~4.8 & ~2.9 & ~0.3 & ~0.9 & ~8.7 & ~7.4 \\
        mT5\textsubscript{1.23B}	 & ~0.3 & ~9.4 & ~0.1 & ~9.9 & 11.1 & ~4.3 & ~3.2 & ~4.4 & ~8.8 & ~3.4 \\
        mT5\textsubscript{3.74B}	 & ~0.2 & ~8.9 & ~0.0 & ~9.0 & ~6.3 & ~2.4 & ~4.3 & ~4.4 & ~7.4 & ~2.6 \\
        mT5\textsubscript{13.0B}	 & ~0.2 & ~8.0 & ~0.1 & ~9.4 & ~8.4 & ~3.0 & ~4.3 & ~4.3 & ~6.8 & ~2.3 \\
        MaLA-500\textsubscript{10.0B}	 & ~0.0 & ~6.7 & ~0.0 & ~7.3 & ~9.7 & ~3.8 & ~0.0 & ~0.0 & ~0.0 & ~0.0 \\
        Teuken Instruct Research\textsubscript{7.0B}	 & ~0.9 & ~6.4 & 12.0 & 46.4 & ~9.7 & ~5.1 & ~2.5 & ~2.5 & 31.2 & 23.0 \\
        Salamandra Instruct\textsubscript{2.0B}	 & ~2.7 & 23.4 & ~3.5 & 34.1 & ~2.0 & ~1.8 & ~0.0 & ~0.0 & ~1.1 & ~0.6 \\
        Salamandra Instruct\textsubscript{7.0B}	 & ~9.1 & 39.3 & ~5.5 & 41.6 & ~2.1 & ~2.6 & ~0.2 & ~0.1 & ~0.0 & ~0.0 \\
        mT0\textsubscript{0.3B}	 & ~4.0 & 17.5 & ~2.2 & 20.0 & 14.3 & 15.3 & ~3.9 & ~1.3 & ~3.3 & ~2.9 \\
        mT0\textsubscript{0.582B}	 & ~0.7 & ~4.8 & ~0.2 & ~5.3 & 21.1 & 21.3 & ~2.6 & ~0.4 & ~4.2 & ~4.9 \\
        mT0\textsubscript{1.23B}	 & ~0.6 & ~6.3 & ~0.2 & ~6.6 & 24.8 & 24.5 & ~5.4 & ~2.4 & ~6.3 & ~6.9 \\
        mT0\textsubscript{3.74B}	 & ~1.7 & 10.0 & ~0.5 & ~8.6 & 30.3 & 29.6 & ~4.7 & ~1.1 & ~9.7 & 13.2 \\
        mT0\textsubscript{13.0B}	 & ~7.8 & 28.3 & ~3.3 & 25.0 & 25.7 & 26.2 & ~3.8 & ~3.2 & 27.4 & 26.8 \\
        \hline
        EuroLLM Instruct\textsubscript{1.7B}	 & ~9.2 & 37.0 & 15.7 & 51.1 & 11.7 & ~5.7 & ~0.0 & ~0.0 & 16.4 & ~9.4 \\
        EuroLLM Instruct\textsubscript{9.0B}	 & ~0.0 & ~0.0 & ~0.0 & ~0.0 & 11.2 & ~4.9 & ~0.0 & ~0.0 & ~0.0 & ~0.0 \\
        Aya-101\textsubscript{12.9B}	 & 26.4 & 56.6 & 19.5 & 52.3 & 36.0 & 32.3 & ~8.2 & ~7.7 & 30.6 & 27.6 \\
        \hline
        Gemma 2\textsubscript{2.0B}	 & ~0.0 & ~0.0 & ~0.0 & ~0.0 & 10.0 & ~4.3 & ~0.2 & ~0.2 & ~0.0 & ~0.0 \\
        Gemma 2\textsubscript{9.0B}	 & ~0.0 & ~0.0 & ~0.0 & ~0.0 & ~9.8 & ~3.9 & ~0.2 & ~0.1 & ~0.0 & ~0.0 \\
        Llama 2\textsubscript{7.0B}	 & ~0.0 & ~0.0 & ~0.0 & ~0.0 & 11.0 & ~4.5 & ~0.0 & ~0.0 & ~0.0 & ~0.0 \\
        Llama 2\textsubscript{13.0B}	 & ~0.0 & ~0.0 & ~0.0 & ~0.0 & 11.3 & ~4.7 & ~0.0 & ~0.0 & ~0.0 & ~0.0 \\
        Llama 3\textsubscript{8.0B}	 & ~3.3 & 12.5 & ~5.3 & 32.8 & 11.3 & ~6.0 & ~0.0 & ~0.0 & ~0.8 & ~0.3 \\
        Ministral Instruct\textsubscript{8.0B}	 & ~1.2 & ~6.7 & ~1.2 & 16.9 & 10.5 & ~5.0 & ~0.2 & ~0.2 & ~0.1 & ~0.0 \\
        Gemma 2 Instruct\textsubscript{2.0B}	 & ~0.0 & ~0.0 & ~0.0 & ~0.0 & 11.6 & ~6.5 & ~0.0 & ~0.0 & ~0.0 & ~0.0 \\
        Gemma 2 Instruct\textsubscript{9.0B}	 & ~0.0 & ~0.0 & ~0.0 & ~0.0 & ~8.0 & ~6.3 & ~0.0 & ~0.0 & ~0.0 & ~0.0 \\
        Llama 2 Chat\textsubscript{7.0B}	 & ~0.0 & ~0.0 & ~0.0 & ~0.0 & 12.3 & ~5.9 & ~0.0 & ~0.0 & ~0.0 & ~0.0 \\
        Llama 2 Chat\textsubscript{13.0B}	 & ~0.0 & ~0.0 & ~0.0 & ~0.0 & 10.0 & ~6.8 & ~0.0 & ~0.0 & ~0.0 & ~0.0 \\
        Llama 3 Instruct\textsubscript{8.0B}	 & ~0.0 & ~2.4 & ~8.8 & 39.9 & 12.6 & ~6.5 & ~1.0 & ~1.1 & 31.7 & 24.1 \\
        \hline
    \end{tabular}
    \caption{Results on \NLG{} tasks for models prompted with English zero-shot instructions.}
    \label{table:results_nlg_0shot}
\end{table*}

\begin{table*}[t]
    \footnotesize
    \setlength{\tabcolsep}{3pt}
    \centering
    \begin{tabular}{|l||c|c|c|c|c|}
        \hline
        \multirow{2}{*}{\textbf{Model}} & \textbf{Sentiment Analysis} & \textbf{SIB-200} & \textbf{Taxi1500} & \textbf{News Categories} & \textbf{MultiEURLEX} \\
         & F1 & F1 & F1 & F1 & F1 \\
        \hline\hline
        PolyLM\textsubscript{1.7B}	 & 52.1 & 19.3 & ~6.3 & ~5.1 & 10.4 \\
        PolyLM\textsubscript{13.0B}	 & 26.0 & 21.6 & ~6.3 & 12.3 & 12.3 \\
        XGLM\textsubscript{0.564B}	 & 48.0 & ~2.2 & ~6.3 & 10.9 & 14.5 \\
        XGLM\textsubscript{1.7B}	 & 50.1 & 15.8 & ~6.3 & 10.4 & 12.5 \\
        XGLM\textsubscript{2.9B}	 & 46.5 & 25.8 & ~6.3 & 13.5 & 13.3 \\
        XGLM\textsubscript{4.5B}	 & 56.4 & 14.2 & ~6.4 & 10.6 & 15.4 \\
        XGLM\textsubscript{7.5B}	 & 32.8 & 26.1 & ~6.3 & ~9.2 & 14.7 \\
        mGPT\textsubscript{1.3B}	 & 39.4 & ~2.2 & ~6.3 & ~3.7 & ~7.3 \\
        mGPT\textsubscript{13.0B}	 & 41.8 & 47.7 & ~6.3 & ~5.1 & ~7.2 \\
        BLOOM\textsubscript{0.56B}	 & 50.0 & ~4.4 & ~6.3 & ~4.2 & 15.9 \\
        BLOOM\textsubscript{2.0B}	 & 41.1 & ~2.2 & ~6.3 & ~5.5 & 14.2 \\
        BLOOM\textsubscript{3.0B}	 & 41.1 & 14.1 & ~6.3 & ~7.2 & 14.0 \\
        BLOOM\textsubscript{8.0B}	 & 54.5 & 41.6 & ~6.3 & ~7.7 & 13.3 \\
        Aya-23\textsubscript{8.0B}	 & 68.3 & 34.6 & ~6.3 & 22.8 & 14.5 \\
        BLOOMZ\textsubscript{0.56B}	 & 51.1 & 34.1 & ~6.3 & ~7.5 & 14.5 \\
        BLOOMZ\textsubscript{2.0B}	 & 60.6 & 23.4 & ~6.3 & 17.8 & 14.7 \\
        BLOOMZ\textsubscript{3.0B}	 & 49.3 & 33.0 & ~6.3 & ~9.6 & 15.1 \\
        BLOOMZ\textsubscript{8.0B}	 & 57.5 & 45.1 & ~6.3 & 12.3 & 13.3 \\
        BX-LLaMA\textsubscript{7.0B}	 & 26.0 & ~7.0 & ~9.0 & ~6.0 & ~6.5 \\
        BX-LLaMA\textsubscript{13.0B}	 & 26.0 & ~2.2 & ~6.3 & ~3.6 & 10.1 \\
        BX-BLOOM\textsubscript{8.0B}	 & 53.2 & 23.8 & ~6.3 & ~4.4 & ~7.5 \\
        \hline
        Salamandra\textsubscript{2.0B}	 & 26.0 & 41.1 & ~6.3 & 14.6 & ~9.2 \\
        Salamandra\textsubscript{7.0B}	 & 79.3 & 48.1 & ~8.3 & 25.9 & 10.8 \\
        EuroLLM\textsubscript{1.7B}	 & 52.2 & 20.9 & ~6.3 & 18.5 & 11.6 \\
        EuroLLM\textsubscript{9.0B}	 & 77.4 & 65.1 & 11.7 & ~8.8 & 10.1 \\
        mT5\textsubscript{0.3B}	 & 26.0 & ~2.4 & ~3.7 & ~5.8 & ~8.3 \\
        mT5\textsubscript{0.582B}	 & 35.5 & ~2.8 & 10.3 & ~4.1 & ~7.2 \\
        mT5\textsubscript{1.23B}	 & 49.6 & ~2.2 & ~6.3 & ~2.1 & 11.6 \\
        mT5\textsubscript{3.74B}	 & 39.4 & ~2.2 & ~6.3 & ~1.2 & ~9.9 \\
        mT5\textsubscript{13.0B}	 & 48.0 & ~2.2 & ~6.4 & ~3.9 & 12.5 \\
        MaLA-500\textsubscript{10.0B}	 & 26.0 & 39.3 & ~6.3 & ~3.2 & ~4.1 \\
        Teuken Instruct Research\textsubscript{7.0B}	 & 76.4 & 56.9 & 22.8 & 17.5 & ~6.8 \\
        Salamandra Instruct\textsubscript{2.0B}	 & 76.1 & 52.4 & ~6.3 & 21.3 & ~8.7 \\
        Salamandra Instruct\textsubscript{7.0B}	 & 65.1 & 64.2 & 17.4 & 24.5 & 12.0 \\
        mT0\textsubscript{0.3B}	 & 54.4 & 15.2 & ~6.3 & 18.9 & 13.7 \\
        mT0\textsubscript{0.582B}	 & 35.1 & 19.5 & ~6.3 & 24.3 & 10.8 \\
        mT0\textsubscript{1.23B}	 & 77.0 & 21.1 & ~6.3 & 18.7 & 15.9 \\
        mT0\textsubscript{3.74B}	 & 28.0 & 30.2 & ~6.3 & 15.4 & 14.5 \\
        mT0\textsubscript{13.0B}	 & 61.1 & 46.4 & ~6.3 & 20.3 & 15.7 \\
        \hline
        EuroLLM Instruct\textsubscript{1.7B}	 & 69.0 & 36.5 & ~6.3 & 24.5 & 10.2 \\
        EuroLLM Instruct\textsubscript{9.0B}	 & 78.9 & 64.7 & 11.7 & 14.8 & ~8.0 \\
        Aya-101\textsubscript{12.9B}	 & 86.5 & 52.3 & 19.7 & 23.5 & 15.2 \\
        \hline
        Gemma 2\textsubscript{2.0B}	 & 30.5 & 36.1 & ~6.3 & 21.6 & 12.1 \\
        Llama 2\textsubscript{7.0B}	 & 36.3 & 31.3 & ~6.3 & 24.1 & 14.8 \\
        Llama 2\textsubscript{13.0B}	 & 60.2 & 47.4 & 10.0 & ~9.7 & 13.6 \\
        Llama 3\textsubscript{8.0B}	 & 80.6 & 42.1 & ~6.3 & 23.0 & 13.2 \\
        Ministral Instruct\textsubscript{8.0B}	 & 70.9 & 43.7 & ~6.3 & ~9.0 & 16.4 \\
        Gemma 2 Instruct\textsubscript{2.0B}	 & 70.2 & 58.3 & 10.1 & ~7.6 & 12.9 \\
        Gemma 2 Instruct\textsubscript{9.0B}	 & 85.0 & 74.3 & 31.9 & 22.6 & ~8.1 \\
        Llama 2 Chat\textsubscript{7.0B}	 & 66.2 & 44.3 & ~8.7 & 24.5 & ~9.5 \\
        Llama 2 Chat\textsubscript{13.0B}	 & 72.2 & 62.6 & 16.9 & 16.5 & 16.0 \\
        Llama 3 Instruct\textsubscript{8.0B}	 & 79.9 & 56.7 & ~8.3 & 18.8 & ~9.7 \\
        \hline
    \end{tabular}
    \caption{Results on \NLU{} tasks for models prompted with English one-shot instructions.}
    \label{table:results_nlu_1shot}
\end{table*}

\begin{table*}[t]
    \footnotesize
    \setlength{\tabcolsep}{3pt}
    \centering
    \begin{tabular}{|l||cc|cc|cc|cc|cc|}
        \hline
        \multirow{2}{*}{\textbf{Model}} & \multicolumn{2}{c|}{\textbf{OPUS-100}} & \multicolumn{2}{c|}{\textbf{Flores-200}} & \multicolumn{2}{c|}{\textbf{WebNLG}} & \multicolumn{2}{c|}{\textbf{EUR-Lex-Sum}} & \multicolumn{2}{c|}{\textbf{News Headlines}} \\
         & BLEU & ChrF & BLEU & ChrF & ChrF & Rouge-L & ChrF & Rouge-L & ChrF & Rouge-L \\
        \hline\hline
        PolyLM\textsubscript{1.7B}	 & ~0.4 & 11.6 & ~0.3 & 16.5 & 10.7 & ~8.9 & ~0.0 & ~0.0 & 11.9 & ~4.2 \\
        PolyLM\textsubscript{13.0B}	 & ~6.5 & 19.1 & ~1.8 & 19.8 & 10.4 & ~6.2 & ~0.0 & ~0.0 & 19.9 & 10.7 \\
        XGLM\textsubscript{0.564B}	 & ~2.8 & 19.0 & ~0.6 & 17.3 & 11.7 & ~6.4 & ~0.0 & ~0.0 & 20.5 & ~9.3 \\
        XGLM\textsubscript{1.7B}	 & ~3.5 & 15.3 & ~1.0 & 17.6 & 13.4 & 10.2 & ~0.0 & ~0.0 & 21.5 & ~9.5 \\
        XGLM\textsubscript{2.9B}	 & ~3.4 & 13.5 & ~1.1 & 13.8 & 12.8 & ~8.6 & ~0.0 & ~0.0 & 21.6 & ~9.8 \\
        XGLM\textsubscript{4.5B}	 & ~3.8 & 15.7 & ~1.1 & 15.6 & 13.9 & ~9.9 & ~0.0 & ~0.0 & 21.5 & ~9.2 \\
        XGLM\textsubscript{7.5B}	 & ~3.2 & 12.7 & ~1.1 & 16.4 & 12.9 & 10.7 & ~0.0 & ~0.0 & 20.8 & ~9.1 \\
        mGPT\textsubscript{1.3B}	 & ~1.6 & 11.6 & ~1.0 & 16.9 & 10.0 & ~6.8 & ~0.2 & ~0.6 & ~1.4 & ~0.7 \\
        mGPT\textsubscript{13.0B}	 & ~1.0 & ~9.2 & ~0.4 & 11.6 & ~9.8 & ~6.1 & ~0.1 & ~0.1 & ~1.0 & ~0.2 \\
        BLOOM\textsubscript{0.56B}	 & ~1.0 & ~9.8 & ~0.4 & 13.2 & 11.2 & ~8.3 & ~0.0 & ~0.0 & 11.0 & ~3.8 \\
        BLOOM\textsubscript{2.0B}	 & ~0.4 & ~9.9 & ~0.2 & 14.5 & 11.6 & ~7.9 & ~0.0 & ~0.0 & 17.8 & ~7.4 \\
        BLOOM\textsubscript{3.0B}	 & ~1.8 & 10.7 & ~0.6 & 14.9 & 12.5 & 10.6 & ~0.0 & ~0.0 & 18.1 & ~7.7 \\
        BLOOM\textsubscript{8.0B}	 & ~1.6 & 11.4 & ~0.9 & 15.1 & 12.7 & 11.2 & ~0.0 & ~0.0 & 21.0 & 10.4 \\
        Aya-23\textsubscript{8.0B}	 & ~2.3 & 13.1 & ~0.5 & 11.6 & 12.3 & ~8.2 & ~0.0 & ~0.0 & 28.2 & 19.8 \\
        BLOOMZ\textsubscript{0.56B}	 & ~0.2 & ~4.6 & ~0.0 & ~5.0 & 10.7 & ~9.7 & ~0.7 & ~1.0 & 11.6 & ~4.1 \\
        BLOOMZ\textsubscript{2.0B}	 & ~1.0 & ~8.8 & ~0.2 & ~9.1 & 12.6 & 12.0 & ~2.0 & ~2.9 & 13.1 & ~4.8 \\
        BLOOMZ\textsubscript{3.0B}	 & ~0.3 & ~5.9 & ~0.2 & ~7.7 & 17.9 & 17.5 & ~0.4 & ~0.5 & 13.2 & ~4.9 \\
        BLOOMZ\textsubscript{8.0B}	 & ~0.5 & ~7.6 & ~0.3 & ~9.3 & 12.6 & 12.8 & ~0.8 & ~1.1 & 12.4 & ~4.5 \\
        BX-LLaMA\textsubscript{7.0B}	 & ~0.0 & ~1.6 & ~0.0 & ~1.4 & ~1.9 & ~0.4 & ~1.2 & ~0.0 & ~0.8 & ~0.0 \\
        BX-LLaMA\textsubscript{13.0B}	 & ~1.0 & 12.0 & ~0.7 & 16.0 & 11.7 & ~7.8 & ~0.1 & ~0.1 & ~8.2 & ~3.2 \\
        BX-BLOOM\textsubscript{8.0B}	 & ~0.6 & 10.3 & ~0.6 & 16.5 & 11.0 & ~9.3 & ~0.0 & ~0.0 & ~5.2 & ~3.8 \\
        \hline
        Salamandra\textsubscript{2.0B}	 & ~3.7 & 22.3 & ~7.1 & 34.3 & ~8.2 & ~6.5 & ~0.0 & ~0.0 & 20.2 & 12.8 \\
        Salamandra\textsubscript{7.0B}	 & ~6.9 & 29.0 & 12.6 & 42.7 & 14.9 & 13.7 & ~0.0 & ~0.0 & 22.8 & 17.9 \\
        EuroLLM\textsubscript{1.7B}	 & 16.4 & 37.6 & 20.9 & 53.0 & 12.4 & ~9.9 & ~0.0 & ~0.0 & 19.4 & 10.5 \\
        EuroLLM\textsubscript{9.0B}	 & 23.6 & 47.1 & 34.0 & 63.7 & 14.4 & 14.1 & ~0.0 & ~0.0 & ~4.4 & ~3.0 \\
        mT5\textsubscript{0.3B}	 & ~0.1 & ~1.8 & ~0.0 & ~2.0 & ~1.5 & ~0.6 & ~0.2 & ~0.5 & ~5.7 & ~4.1 \\
        mT5\textsubscript{0.582B}	 & ~0.4 & ~4.7 & ~0.1 & ~5.9 & ~6.1 & ~7.0 & ~0.3 & ~0.9 & ~5.9 & ~5.4 \\
        mT5\textsubscript{1.23B}	 & ~0.2 & ~9.0 & ~0.1 & 12.4 & 10.6 & ~3.4 & ~3.2 & ~4.4 & ~6.2 & ~1.7 \\
        mT5\textsubscript{3.74B}	 & ~0.3 & 11.0 & ~0.2 & 15.1 & ~8.9 & ~3.0 & ~4.3 & ~4.4 & ~5.7 & ~1.6 \\
        mT5\textsubscript{13.0B}	 & ~0.2 & ~8.2 & ~0.1 & 11.0 & ~9.0 & ~3.1 & ~4.3 & ~4.4 & ~8.7 & ~3.4 \\
        MaLA-500\textsubscript{10.0B}	 & ~0.0 & 11.2 & ~0.1 & 13.7 & ~9.8 & ~5.5 & ~0.0 & ~0.0 & 13.2 & ~4.8 \\
        Teuken Instruct Research\textsubscript{7.0B}	 & 18.8 & 45.8 & 18.8 & 54.7 & 13.6 & 14.1 & ~2.5 & ~2.4 & 31.0 & 24.1 \\
        Salamandra Instruct\textsubscript{2.0B}	 & ~4.2 & 28.1 & ~3.4 & 35.2 & ~3.3 & ~1.9 & ~0.0 & ~0.0 & 18.9 & 12.7 \\
        Salamandra Instruct\textsubscript{7.0B}	 & ~8.5 & 43.4 & ~6.2 & 44.3 & 13.2 & 10.3 & ~0.2 & ~0.1 & 25.7 & 18.9 \\
        mT0\textsubscript{0.3B}	 & ~2.5 & 16.8 & ~0.6 & 14.5 & ~5.0 & ~3.9 & ~3.9 & ~1.3 & ~0.6 & ~0.1 \\
        mT0\textsubscript{0.582B}	 & ~0.4 & ~5.7 & ~0.0 & ~3.9 & 13.3 & ~9.2 & ~2.6 & ~0.4 & ~8.4 & ~2.3 \\
        mT0\textsubscript{1.23B}	 & ~1.1 & 10.5 & ~0.2 & 10.2 & 13.8 & 10.8 & ~5.4 & ~2.4 & ~1.8 & ~1.8 \\
        mT0\textsubscript{3.74B}	 & ~1.4 & 10.3 & ~0.1 & ~7.1 & 18.1 & 17.1 & ~4.7 & ~1.1 & 12.3 & ~4.5 \\
        mT0\textsubscript{13.0B}	 & ~9.0 & 31.3 & ~4.8 & 29.6 & 10.8 & 10.8 & ~3.8 & ~3.2 & 11.6 & ~6.5 \\
        \hline
        EuroLLM Instruct\textsubscript{1.7B}	 & 13.1 & 38.3 & 21.0 & 54.0 & 13.3 & 10.5 & ~0.0 & ~0.0 & 22.8 & 14.3 \\
        EuroLLM Instruct\textsubscript{9.0B}	 & 33.6 & 60.4 & 38.6 & 68.7 & 17.5 & 18.2 & ~0.0 & ~0.0 & 35.2 & 28.0 \\
        Aya-101\textsubscript{12.9B}	 & 26.8 & 57.8 & 21.2 & 54.3 & 41.2 & 36.4 & ~8.2 & ~7.7 & 30.0 & 28.0 \\
        \hline
        Gemma 2\textsubscript{2.0B}	 & ~3.3 & 14.3 & ~2.4 & 23.7 & 13.0 & 13.1 & ~0.2 & ~0.2 & 19.2 & 12.6 \\
        Llama 2\textsubscript{7.0B}	 & ~4.6 & 17.0 & ~1.5 & 19.3 & 12.6 & 10.3 & ~0.0 & ~0.0 & 23.0 & 16.0 \\
        Llama 2\textsubscript{13.0B}	 & ~6.3 & 23.1 & ~2.0 & 23.5 & 16.7 & 15.9 & ~0.0 & ~0.0 & 26.7 & 19.4 \\
        Llama 3\textsubscript{8.0B}	 & 14.9 & 39.7 & 10.1 & 41.8 & 17.4 & 17.8 & ~0.0 & ~0.0 & 33.0 & 25.2 \\
        Ministral Instruct\textsubscript{8.0B}	 & ~3.9 & 17.2 & ~1.4 & 18.8 & 14.4 & 11.9 & ~0.8 & ~1.5 & 30.9 & 21.3 \\
        Gemma 2 Instruct\textsubscript{2.0B}	 & ~3.1 & 13.6 & ~1.7 & 19.5 & 12.2 & ~8.9 & ~0.0 & ~0.0 & ~9.8 & ~6.5 \\
        Gemma 2 Instruct\textsubscript{9.0B}	 & 14.4 & 36.7 & 18.4 & 52.4 & 22.8 & 23.3 & ~0.0 & ~0.0 & 32.3 & 27.0 \\
        Llama 2 Chat\textsubscript{7.0B}	 & ~3.8 & 16.3 & ~2.5 & 20.0 & 14.0 & 13.4 & ~0.0 & ~0.0 & 25.7 & 18.4 \\
        Llama 2 Chat\textsubscript{13.0B}	 & ~7.9 & 25.8 & ~4.0 & 27.6 & 15.2 & 15.5 & ~0.0 & ~0.0 & 29.4 & 20.8 \\
        Llama 3 Instruct\textsubscript{8.0B}	 & 17.6 & 44.7 & 10.9 & 43.7 & 30.3 & 26.4 & ~1.0 & ~1.1 & 33.6 & 26.6 \\
        \hline
    \end{tabular}
    \caption{Results on \NLG{} tasks for models prompted with English one-shot instructions.}
    \label{table:results_nlg_1shot}
\end{table*}

\begin{table*}[t]
    \centering
    \begin{tabular}{|l||c|c|c|c|c|}
        \hline
        \multirow{2}{*}{\textbf{Model}} & \textbf{Sentiment} & \textbf{SIB-200} & \textbf{Taxi1500} & \textbf{News Categories} & \textbf{MultiEURLEX} \\
        & Macro-F1 & Macro-F1 & Macro-F1 & Macro-F1 & Macro-F1 \\
        \hline\hline
        BERTu & 83.0 & 84.9 & 77.5 & 58.3 & 67.1 \\
        mBERT & 64.7 & 75.3 & 47.7 & 53.1 & 60.7 \\
        XLM-R & 59.6 & 68.5 & 36.5 & 50.6 & 60.5 \\
        Glot500 & 74.6 & 82.3 & 64.0 & 57.2 & 62.2 \\
        mT5 & 100.0 & 76.8 & 42.2 & 52.5 & 31.2 \\
        \hline
    \end{tabular}
    \caption{Fine-tuned model results on \NLU{} tasks.}
    \label{table:results_nlu_finetuned}
\end{table*}

\begin{table*}[t]
    \centering
    \begin{tabular}{|l||cc|cc|cc|cc|}
        \hline
        \multirow{2}{*}{\textbf{Model}} & \multicolumn{2}{c|}{\textbf{OPUS-100}} & \multicolumn{2}{c|}{\textbf{WebNLG}} & \multicolumn{2}{c|}{\textbf{EUR-Lex-Sum}} & \multicolumn{2}{c|}{\textbf{News Headlines}} \\
        & BLEU & ChrF & ChrF & Rouge-L & ChrF & Rouge-L & ChrF & Rouge-L \\
        \hline\hline
        mT5 (fine-tuned) & 51.8 & 75.9 & 31.6 & 28.0 & 51.5 & 42.5 & 32.2 & 28.1 \\
        \hline
    \end{tabular}
    \caption{Fine-tuned model results on \NLG{} tasks.}
    \label{table:results_nlg_finetuned}
\end{table*}

\section{Closed-Source Model Results}
\label{appendix:chatgpt_result}

We include experiments with ChatGPT 4o~\cite{openai-2024-gpt4} as a comparison to our main experiments.
However, since this is a closed-source model accessible only through an API, our experiments with this model are limited.
Firstly, since we do not have access to log-likelihoods, it is not possible for us to conduct \NLU{} experiments in a comparative manner, so we skip these tasks.
Secondly, we also skip EUR-Lex-Sum due to the large context lengths needed for this task, which exceed our quota.
Thirdly, for the remaining four tasks, we only prompt with 100 test samples for each task to limit our costs.

\begin{table*}[t]
    \centering
    \begin{tabular}{|ll||cc|cc|cc|cc|}
        \hline
        \multirow{2}{*}{\textbf{Prompt}} & \multirow{2}{*}{\textbf{Shots}} & \multicolumn{2}{c|}{\textbf{OPUS-100}} & \multicolumn{2}{c|}{\textbf{Flores-200}} & \multicolumn{2}{c|}{\textbf{WebNLG}} & \multicolumn{2}{c|}{\textbf{News Headlines}} \\
        & & BLEU & ChrF & BLEU & ChrF & ChrF & Rouge-L & ChrF & Rouge-L \\
        \hline\hline
        English & Zero & 38.1 & 69.7 & 44.1 & 74.4 & 61.8 & 58.1 & 32.1 & 26.6 \\
        Maltese & Zero &34.2 & 64.0 & 43.5 & 72.4 & 56.0 & 53.6 & 32.2 & 27.2 \\
        \hline
        English & One & 36.8 & 67.3 & 46.2 & 74.5 & 61.9 & 57.4 & 33.8 & 25.9 \\
        Maltese & One & 35.8 & 65.6 & 46.2 & 74.5 & 61.8 & 58.3 & 31.9 & 24.7 \\
        \hline
    \end{tabular}
    \caption{ChatGPT Results.}
    \label{table:results_chatgpt}
\end{table*}

The results are shown in Table~\ref{table:results_chatgpt}.
Comparing these figures to the results obtained by our fine-tuned mT5 baseline (Table~\ref{table:results_nlg_finetuned}), ChatGPT 4o performs significantly worse on OPUS-100, significantly better on WebNLG and on par on Maltese News Headlines.

\section{Analysing Other Model Properties}
\label{appendix:other_model_properties}

\subsection{Model Size}
\label{appendix:model_size}

We look at the performance as the model size grows in terms of the number of parameters.
To analyse this, we fit linear regression models for \PT{} and \IT{} models on performance results aggregated by task type.
We only do this for zero-shot results.

\begin{figure*}[t]
\begin{subfigure}{0.5\linewidth}
    \centering
    \includegraphics[scale=0.8]{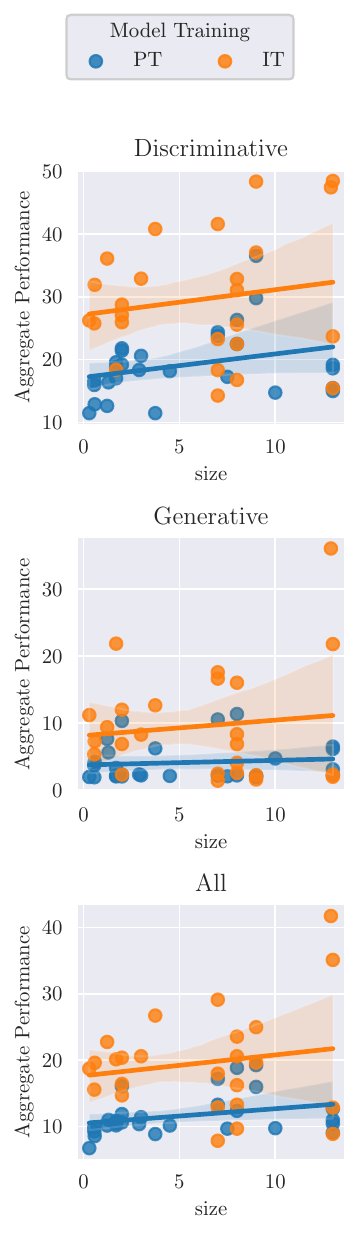}
    \caption{Including \PT{}/\PT{}, \IT{}/\PT{}, and \IT{}/\IT{} models}
    \label{figure:model_size_0shot_performance}
\end{subfigure}
\begin{subfigure}{0.5\linewidth}
    \centering
    \includegraphics[scale=0.8]{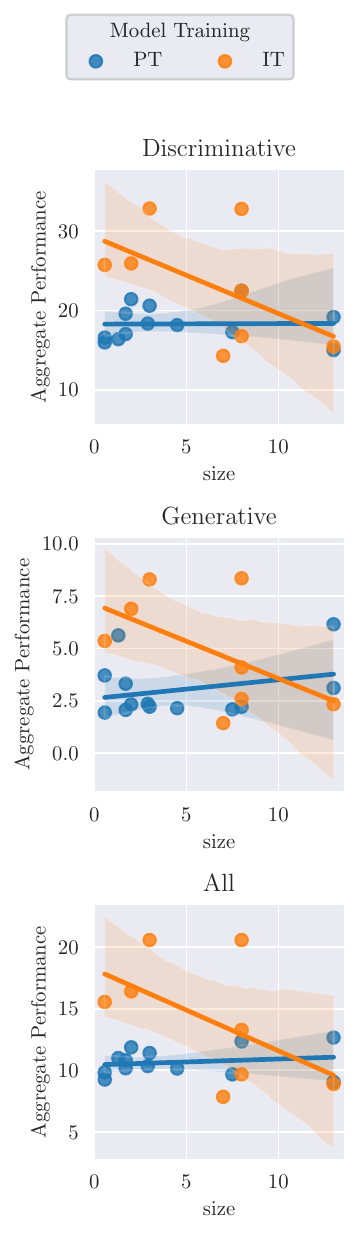}
    \caption{Excluding \PT{}/\PT{}, \IT{}/\PT{}, and \IT{}/\IT{} models}
    \label{figure:model_size_0shot_performance_no_maltese}
\end{subfigure}
\caption{Zero-shot aggregated performance against model size.}
\end{figure*}

As shown in Figure~\ref{figure:model_size_0shot_performance}, a general improvement is observed in performance with model size increase.
In general, \IT{} models with larger sizes give better performances than \PT{} models.
In fact, a smaller performance gap is observed between \PT{} and \IT{} models with fewer than 10B parameters, especially in few-shot, where \PT{} models overall perform better than \IT{} models on \NLG{} tasks.

However, we note that among the largest \IT{} models are the Aya-101 and mT0 models, which are trained on Maltese.
If we exclude models which we know are trained on Maltese, then our previous observations do not hold as shown in Figure~\ref{figure:model_size_0shot_performance_no_maltese}.
In fact, we see a negative impact on performance as model size grows for \IT{} models, albeit with a larger confidence interval.

\subsection{Model Multilinguality}
\label{appendix:model_multilinguality}

We also analysed a model's performance against the number of languages it was exposed to during its training.
Similar to Section~\ref{section:maltese_analysis}, we exclude models with unknown Maltese training (\NK{}).
We also exclude MaLA-500 from this analysis, as the high degree of languages skews our plots.
Other than that, we plot aggregated zero-shot performance results against model multilinguality and fit separate linear regression models for \PT{} and \IT{} models.
 
\begin{figure*}[t]
\begin{subfigure}{0.5\linewidth}
    \centering
    \includegraphics[scale=0.8]{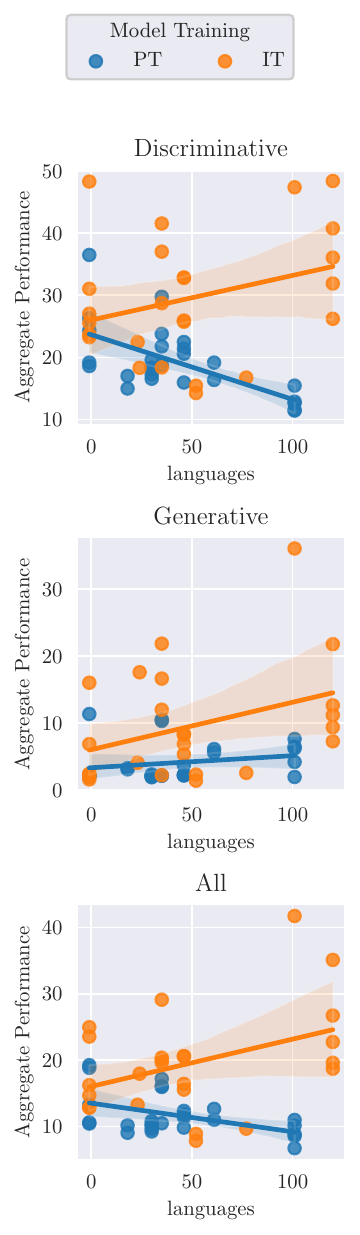}
    \caption{Including \PT{}/\PT{}, \IT{}/\PT{}, and \IT{}/\IT{} models}
    \label{figure:model_multilinguality_0shot_performance}
\end{subfigure}
\begin{subfigure}{0.5\linewidth}
    \centering
    \includegraphics[scale=0.8]{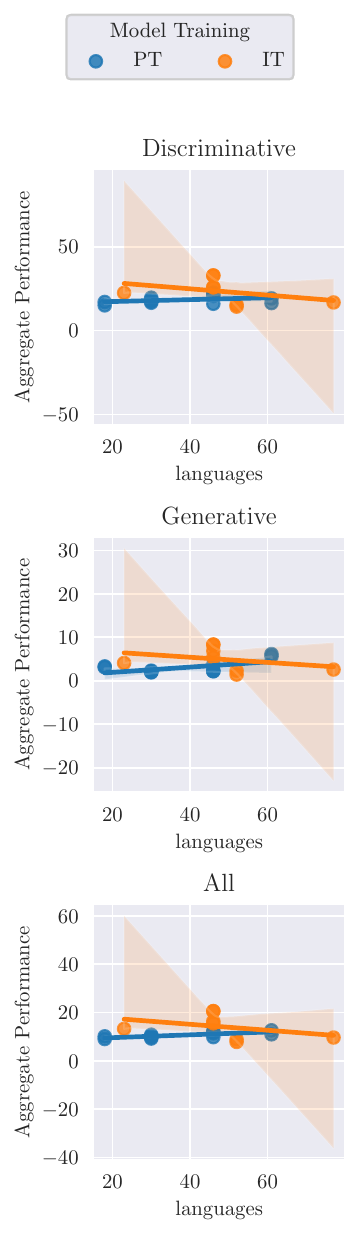}
    \caption{Excluding \PT{}/\PT{}, \IT{}/\PT{}, and \IT{}/\IT{} models}
    \label{figure:model_multilinguality_0shot_performance_no_maltese}
\end{subfigure}
\caption{Zero-shot aggregated performance against model multilinguality.}
\end{figure*}

In Figure~\ref{figure:model_multilinguality_0shot_performance} we observe a positive influence with the number of languages a model is exposed to for \IT{} models.
For \PT{} models there is also a positive effect on \NLG{} tasks, although smaller than that for \IT{} models.
On the other hand, there is a negative impact as the number of languages increases for \NLU{} tasks.

Despite this, highly multilingual models which have seen more than 100 languages, are all models which have been exposed to Maltese during some part of their training.
When excluding these models and refitting logistic models on the remaining data we see that the previously observed improvements are drastically reduced.

\end{document}